\documentclass{article}

\usepackage{microtype}
\usepackage{graphicx}   \usepackage{subcaption} \usepackage{booktabs}   \usepackage{amsmath}
\usepackage{amssymb}
\usepackage{xcolor}
\usepackage{tcolorbox}
\usepackage{tikz}       \usetikzlibrary{arrows.meta, positioning, fit, patterns, shapes.geometric}
\usepackage{paralist}
\usepackage{enumitem}
\usepackage{outlines}
\usepackage[small, compact]{titlesec}
\titlespacing{\paragraph}{0pt}{-\parskip}{-\parskip}

\usepackage{hyperref}

\newcommand{\eat}[1]{}

\usepackage[accepted]{mlsys2025}
\makeatletter
\renewcommand{\Notice@String}{Preprint. Under review.}
\makeatother

\usepackage{cleveref}  \crefformat{section}{\S#2#1#3}
\Crefformat{section}{\S#2#1#3}
\crefformat{subsection}{\S#2#1#3}
\Crefformat{subsection}{\S#2#1#3}

\mlsystitlerunning{Maximizing GPU Utilization in On-Policy RL}

\newif\ifdraft
\draftfalse   

\ifdraft
    \newcommand{\mat}[1]{{\color{blue}[Mat: #1]}}  \newcommand{\mbw}[1]{\textcolor{olive}{[MBW: #1]}}
    
    \newcommand{\swang}[1]{\textcolor{purple}{[Swang: #1]}}
    
    \newcommand{\dr}[1]{\textcolor{magenta}{[DR: #1]}}
    
    \newcommand{\kan}[1]{\textcolor{brown}{[Kan: #1]}}
    \newcommand{\yi}[1]{\textcolor{pink}{[Yi: #1]}}
    \newcommand{\lcz}[1]{\textcolor{orange}{[Luis: #1]}}

\else
    \newcommand{\mat}[1]{}
    \newcommand{\mbw}[1]{}
    \newcommand{\swang}[1]{}
    \newcommand{\dr}[1]{}
    \newcommand{\kan}[1]{}
    \newcommand{\yi}[1]{}
    \newcommand{\lcz}[1]{}
    
\fi

\newcommand{\RR}[0]{Rollout Router}
\newcommand{\TR}[0]{Train Router}

\definecolor{gapred}{HTML}{D55E00}
\definecolor{takeblue}{HTML}{0072B2}
\newtcolorbox{motivbox}[1]{colback=#1!6, colframe=#1, boxrule=0pt, leftrule=2.5pt,
  arc=0pt, outer arc=0pt, left=5pt, right=5pt, top=3pt, bottom=3pt, before skip=0.6em, after skip=0.6em}

\usepackage{xspace}
\newcommand{\sys}{\textsc{TIDE}\xspace}
\newcommand{\sysm}{\textsc{TIDE-Megatron}\xspace}
\newcommand{\sysf}{\textsc{TIDE-FSDP}\xspace}

\begin{document}

\twocolumn[
\mlsystitle{TIDE: Reclaiming Lost Performance in On-Policy Reinforcement Learning}

\begin{mlsysauthorlist}
\mlsysauthor{Mathew Jacob\textsuperscript{*}}{uw}
\mlsysauthor{Morgan Borjigin-Wang\textsuperscript{*}}{brown}
\mlsysauthor{Kan Zhu}{uw}
\mlsysauthor{Po Hao Chen}{brown}
\mlsysauthor{Yi Pan}{berkeley}
\par
\mlsysauthor{Andy Ruan}{uw}
\mlsysauthor{Nayani Modugula}{nyu}
\mlsysauthor{Ethan Lavi}{brown}
\mlsysauthor{Luis Ceze}{uw}
\mlsysauthor{Malte Schwarzkopf}{brown}
\par
\mlsysauthor{Akshay Narayan}{brown}
\mlsysauthor{Deepti Raghavan}{brown}
\mlsysauthor{Stephanie Wang}{uw}
\end{mlsysauthorlist}

\mlsysaffiliation{uw}{University of Washington, Seattle, Washington, USA}
\mlsysaffiliation{brown}{Brown University, Providence, Rhode Island, USA}
\mlsysaffiliation{berkeley}{UC Berkeley, Berkeley, California, USA}
\mlsysaffiliation{nyu}{New York University, New York, New York, USA}

\mlsyscorrespondingauthor{Mathew Jacob}{mjacob2@cs.washington.edu}
\mlsyscorrespondingauthor{Morgan Borjigin-Wang}{\mbox{mbwang@brown.edu}}

\vskip 0.2in

\begin{abstract}
Synchronous, colocated RL post-training of large language models (LLMs) today suffers from severe under-utilization due to \emph{long-tail rollouts}.
Inference engines fall idle waiting for long-tail rollouts or decode at low, inefficient batch sizes.
To address this, we present \sys, an RL training system that preserves on-policy training semantics while recovering resource utilization.
\sys{} does so by two complementary techniques: (1) dynamically repurposing rollout GPUs for training while long-tail rollouts complete and (2) dynamically migrating rollouts between inference engines to boost decoding throughput.
\sys automatically determines the right time to repurpose rollout resources through an adaptive migration policy, obtaining performance improvements that generalize across multiple applications and hardware configurations without requiring bespoke configuration.
In our evaluation, \sys provides end-to-end speedups of up to $1.36\times$ over colocated systems, outperforming all evaluated on-policy baselines and achieving up to 97\% of optimal throughput.
\end{abstract}
]
\printAffiliationsAndNotice{\mlsysEqualContribution}

\section{Introduction}

Reinforcement learning (RL) is now an essential step in post-training of large language models (LLMs).
RL techniques adapt
models to specialized domains, improve general model reasoning capabilities, and align model outputs with human preferences~\cite{DeepseekR1, openaio1, olmo3}.
RL works by repeatedly executing \emph{rollouts} to generate data online using the current model weights, then \emph{trains} on the generated data to update the model.
The rollout and training phases have different computational characteristics and each employs specialized systems, such as vLLM~\cite{vllm} or SGLang~\cite{sglang} for inference and Megatron-LM~\cite{Megatron} or PyTorch FSDP~\cite{fsdp} for training. To make efficient use of GPUs, many RL systems \emph{colocate} both phases on the same hardware, switching between inference and training systems within each RL step.

However, this colocated setup results in significant resource under-utilization.
Current implementations are \emph{synchronous}: they insert a barrier between inference and training so that all GPUs first execute rollouts, then after all rollouts complete, the entire cluster proceeds to the training phase.
This is a simple way to ensure \emph{on-policy training}, which requires that the training inputs come from current model weights.

Unfortunately, the barrier also creates two sources of under-utilization.
First, due to the variation in rollout completion times, a single straggler will cause all other GPUs to remain \emph{idle} in the rollout phase. Moreover, even the GPUs that remain active lose \emph{decoding efficiency} as rollouts complete and decode batch sizes shrink. 
Across two different RL tasks on GPU configurations ranging from 4×L40S to 8×H200, we find that GPU idleness accounts for up to 30.0\% of total run time, with inefficient decoding contributing another 9.8\%. 
Together with phase-switching overheads (0.2-3.6\%), these losses leave conventional colocated execution achieving as little as 56.6\% of the optimal utilization.

Some prior efforts to address these inefficiencies adopt asynchronous, off-policy RL training~\cite{AReaL}.
While this allows for higher GPU utilization, off-policy training must use stale weights and thus changes the training semantics.
Other efforts preserve on-policy training~\cite{hu2026tlt} but empirically fail to generalize to RL training's diverse workloads~(\Cref{sec:evaluation}), which can span many end tasks, training backends, and hardware topologies.

We therefore ask: \textit{can we eliminate the underutilization in colocated RL while preserving on-policy training semantics, without requiring workload-specific tuning?}

Our key observation is that \emph{preserving on-policy training requires a barrier only before the weight update, not before training begins}. Although the policy must remain fixed while generating rollouts for the current step, training can begin on completed rollouts while others are still being generated, provided that the weight update is deferred until training work is complete.
Deferring the traditional barrier between rollout and training to before the weight update creates an opportunity to overlap inference and training within one RL step, using GPU capacity for training that would otherwise remain idle.

\begin{figure}[t]
    \centering
    \includegraphics[width=\columnwidth]{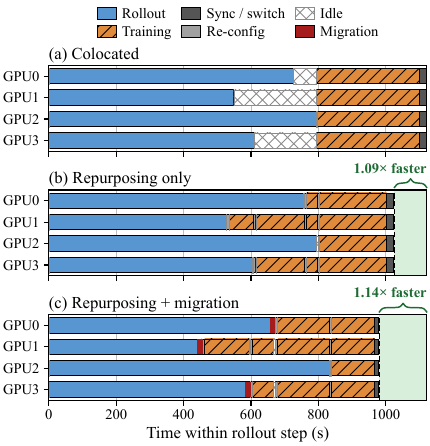}
    \caption{DAPO Math on 4$\times$H100. Repurposing fills the idle time of the
  colocated baseline, and migration raises decode throughput through larger
  batches, together reducing the GPU-hours spent in the rollout phase.}
    \label{fig:system_timeline}
\end{figure}
 
Building on this observation, we introduce \sys{}, an on-policy RL training  system that uses fine-grained resource reallocation to maximize GPU utilization. 
\sys builds on two complementary mechanisms.
First, we \emph{repurpose} GPUs that have finished their rollouts to begin training on completed rollouts, enabling overlap of the rollout and training phases. 
Second, \emph{request migration} consolidates outstanding rollouts onto fewer GPUs to improve decoding efficiency per GPU, while enabling further GPU repurposing.
Together, these mechanisms allow \sys{} to reclaim lost GPU-hours from both idleness and decoding inefficiency.
We illustrate these components of our system in Figure \ref{fig:system_timeline}.

A key challenge in \sys is determining the best time during the rollout phase to migrate rollouts and repurpose GPUs to training.
Both repurposing too soon and too late lead to under-utilization, and the best time depends on application characteristics and hardware configuration (\S\ref{sec:performance_analysis}).
To address this, we design a robust migration policy that uses a simple and instantaneous signal to estimate the balance between rollout vs.~training throughput, combined with a linear controller inspired by network congestion control.

A second challenge in \sys{} is handling the now dynamically growing training group, while preserving on-policy semantics and supporting diverse training backends.
We design an orchestrator for the training group to dynamically assign completed rollouts and manage synchronization across training GPUs.
We also show that \sys{}'s techniques generalize across training backends, both data parallelism (DP) with tensor parallelism (TP) via Megatron and ZeRO via PyTorch FSDP.

Overall, we show that \sys consistently achieves the best speedup over a colocated baseline compared to prior efforts that preserve on-policy training but fail to fully capture GPU waste.
Across a range of workloads and hardware, \sys{} achieves up to 36\% faster runtime than the colocated baseline.
Furthermore, \sys's repurposing and migration policies effectively recover 77-97\% of the optimal possible throughput.

In summary, this paper makes the following contributions:
\begin{itemize}[noitemsep, topsep=0pt]
    \item We systematically categorize and quantify the sources of under-utilization in synchronous colocated RL training and identify the missed opportunities (\S\ref{sec:performance_analysis}).
    \item We design an adaptive policy for repurposing under-utilized rollout GPUs (\S\ref{s:migration_policy}) and build a system, \sys, that generalizes to a range of application, training backends, and hardware.
\end{itemize}

 \section{Related Work}
\label{sec:related-work}

\paragraph{Overlapping-Based Methods.}
Like \sys, some past RL systems attempt to repurpose GPUs to overlap tasks and hide latency across or within stages of the RL pipeline.

RollPacker~\cite{RollPacker} proposes tail batching: it over-provisions rollout samples and trains on the fastest completions.
The unfinished long tails are batched into a later training step. Tail batching implicitly skews the distribution of the training samples, so it does not preserve the original RL semantics.
RollPacker additionally proposes overlapping inference and training by opportunistically switching a subset of GPUs from inference to training using a policy that switches exactly once when certain heuristic conditions are satisfied. 
\sys repurposes GPUs at a finer granularity, and we show in \S\ref{s:eval:e2e} that this results in greater speedups.

RLHFuse~\cite{RLHFuse} overlaps rollout generation with reward computation, but not training.
RLHFuse also focuses on the PPO algorithm~\cite{PPO}, and does not apply to the critic-free GRPO~\cite{GRPO} setting that is popular today and which \sys targets.

\paragraph{Speeding Up Long-Tail Inference.}
RL systems like TLT~\cite{hu2026tlt}, DAS~\cite{shao2026beatlongtail}, and ReSpec~\cite{chen2026respec} 
attempt to mitigate straggler rollouts via speculative decoding (SD).
Other methods attempt to use previous historical rollouts of the same prompt across epochs as a speculative draft for the rollout in a future RL step\cite{RhymeRL}.

In \S\ref{s:eval:e2e}, we show that \sys outperforms TLT due to the implicit tradeoffs of enabling SD. 
Further, SD cannot accelerate tool calls and relies on training an acceptable draft model for good performance. 

\paragraph{Training-Inference Disaggregation.}
Instead of a colocated architecture, some RL systems 
such as StreamRL~\cite{StreamRL} and ThunderSyncRL~\cite{ThunderSyncRL}
instead 
partition GPUs into separate inference and training pools and stream completed rollouts to training engines.
This allows, for example, using hardware specialized for inference and training, respectively.
However, disaggregation 
\begin{inparaenum}[(i)]
    \item requires significantly more GPU resources, and
    \item requires careful tuning to balance inference and training resource allocations.
\end{inparaenum}
\sys similarly streams generated samples to trainers, but 
does so for colocated architectures and automatically balances inference and training resource allocations.

\paragraph{Algorithmic Modifications.}
Asynchronous RL systems such as AReaL~\cite{AReaL} make algorithmic modifications to the RL workflow by allowing off-policy rollouts, i.e., rollouts from older iterations of the model.
While this design supports high GPU utilization, training on rollouts from prior model versions changes the training semantics and can result in training instability. 
More recently, QwenGyre~\cite{QwenGyre} also allocates GPUs elastically between inference and training, but uses off-policy learning and requires a static configuration of its repurposing threshold.
While \sys focuses on improving utilization while maintaining the semantics of synchronous on-policy RL training, its techniques---specifically, migrating rollouts and automatically determining the migration threshold---could apply to repurposing GPUs in other RL training designs as well.

 \section{Addressing Under-Utilization}
\label{sec:performance_analysis}

This section quantifies the impact of two sources of GPU under-utilization, both caused by long-tailed rollouts (\S\ref{subsec:gpu_underutilization}).
First, GPUs that finish their work early sit idle waiting for the last GPU.
Second, as each inference engine nears the end of its assigned work, its batch size decreases, causing under-utilization.
We then quantify the potential speedup from repurposing GPUs mid-step (\S\ref{subsec:opportunity}) and the challenges \sys must solve to realize this speedup (\S\ref{subsec:challenges}).

\subsection{Sources of GPU Under-utilization }
\label{subsec:gpu_underutilization}

\begin{figure}[t]
\centering
\includegraphics[width=\linewidth]{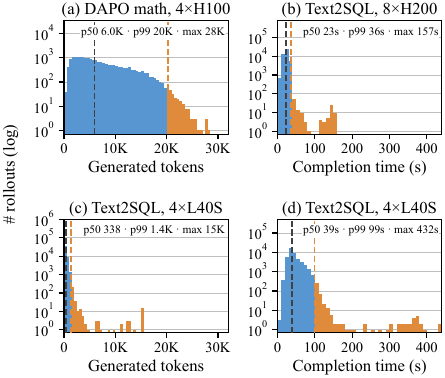}
\caption{Rollout lengths vary across workloads (a, c), and completion times
  vary across hardware (b, d).}
\label{fig:rollout_tail_distribution}
\end{figure}

\paragraph{Long-Tailed Rollouts.}
\label{sss:long_tailed_rollouts}
In RL rollouts, 
stragglers can arise from long responses, multi-turn rollouts, and long tool calls. 
Stragglers' occurrence varies across workloads, models, and hardware.
For example, \Cref{fig:rollout_tail_distribution} (a) and (c) show the variance in sequence length distribution across two datasets (\S\ref{s:eval:setup:apps} describes these datasets in detail).
DAPO Math has a p99 response length of $20$k tokens and a maximum response length of $28$k, while Text2SQL, which requires multi-turn generation and tool use, has a p99 and max response length of $1.4$k and $15$k, respectively.

Hardware differences also influence stragglers.
For example, \Cref{fig:rollout_tail_distribution} (b) and (d) show rollout completion times on NVIDIA 8xH200 and 4xL40S, respectively.
For Text2SQL, running the benchmark on 8xH200s results in a p99 and maximum completion time of $36$s and $157$s, respectively;
on the 4xL40S, the p99 and maximum completion time of rollouts is $99$s and $432$s, respectively.

\begin{figure}[t]
\centering
\includegraphics[width=\linewidth]{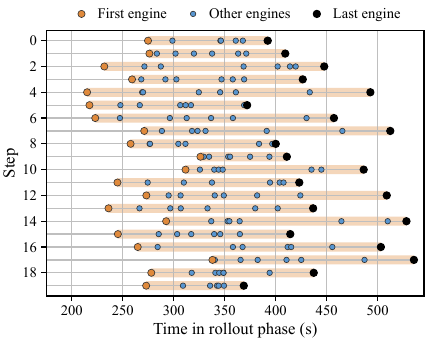}
\caption{Skew in inference-engine finish times across a 20-step colocated
  DAPO-Math run (DeepSeek-R1-Distill-8B, 8$\times$H200). In every step,
  most engines finish well before the last one, leaving their GPUs idle.
  }
\label{fig:idle_skew_figure}
\end{figure}

\paragraph{Idle GPUs} 
\label{sss:idle_gpus}
When stragglers occur, 
some inference engines finish the rollout stage long after the rest (\Cref{fig:idle_skew_figure}).
With a barrier between the rollout and training phases, all inference engines must complete before training begins.
So, while engines processing stragglers complete, other inference engines \emph{idle}, wasting up to 30\% of GPU-hours.

 \paragraph{Decode Inefficiency} 
\label{sss:inefficient_inference}
\begin{figure}[t]
\centering
\includegraphics[width=\linewidth]{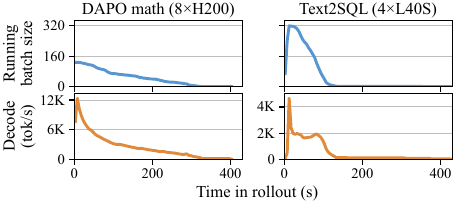}
\caption{As rollouts complete, the batch shrinks and throughput falls far below
  peak while engines decode the last few stragglers.}
\label{fig:batch_size_over_time}
\end{figure}
 
Stragglers can cause under-utilization even for non-idle GPUs.
While training is typically compute-bound, LLM decoding is memory-bound since generating a single token requires loading the full model weights.
Batching multiple sequences together amortizes this load.
However, on-policy RL uses a fixed set of rollouts in the inference phase, so as \Cref{fig:batch_size_over_time} shows, running batch sizes 
and correspondingly decode throughput per engine declines over time, resulting in wasted GPU-hours from underutilized hardware. 

To quantify this \textit{decode inefficiency loss}, we measure the throughput of a single GPU running all of the rollouts for a step. This allows the batch size on the GPU to be as large as possible, for as long as possible, maximizing the throughput and minimizing the number of GPU-hours the rollout stage takes.
This allows us to measure the best-case time required to generate all tokens in a step.

We then measure the GPU-hours spent by the colocated RL system during active rollout generation and compare it with the maximal throughput's GPU hours. 
We see that during inference, on the DAPOxH200 configuration, colocated achieves only 84\% of optimal decode throughput, resulting in 12.6\% wasted rollout-GPU hours, or 8.4\% wasted GPU hours across the entire run.

\begin{table}[t]
\centering
\caption{Breakdown of each step's GPU hours under colocated execution (\%).
  The three losses and the fraction of optimal throughput sum to 100\%.
  Misc.\ idle is GPU time outside both phases, spent offloading and
  onloading engine and trainer state at each phase switch.}
\label{tab:optimality_table}
\vskip 0.1in
\small
\setlength{\tabcolsep}{3pt}
\begin{tabular*}{\columnwidth}{@{\extracolsep{\fill}}llcccc@{}}
  \toprule
                         &               & Idle & Ineff. & Misc. & Of               \\
  Application            & GPUs          & loss & loss   & idle  & optimal          \\
  \midrule
  DAPO Math              & 8$\times$H200 & 15.2 & 8.4    & 1.3   & \textbf{75.1}    \\
  \quad R1-Distill-8B    & 4$\times$H100 & 12.7 & 3.9    & 0.2   & \textbf{83.2}    \\
\addlinespace[2pt]
  Text2SQL               & 8$\times$H200 & 30.0 & 9.8    & 3.6   & \textbf{56.6}    \\
  \quad Qwen2.5-Coder-7B & 4$\times$L40S & 18.5 & 2.5    & 0.6   & \textbf{78.4}    \\
\bottomrule
\end{tabular*}
\end{table} \paragraph{Utilization-optimal}
We now seek to understand the optimal performance on-policy RL can achieve on a given workload, if GPU utilization is maximized.
We assume that because training is compute-bound, it already operates at its peak throughput.
Thus, we can compute the optimal GPU-hours by adding the optimal inference GPU-time described above with the GPU-time needed for training, collected from the actual workload.
We also add the actual time needed to weight-sync between the inference and training engines.
We call this upper bound the overall utilization-optimal.

The percentage of utilization-optimal that synchronous colocated RL achieves under various workloads and settings is recorded in Table \ref{tab:optimality_table}. 
Across both hardware setups for the Text2SQL and DAPO Math datasets, idle and inefficient GPU time together cost $17$--$40\%$ of the GPU hours in each step (Table \ref{tab:optimality_table}).
Assuming that the training phase is already fully utilized, colocated execution thus achieves only $57$--$83\%$ of the optimal throughput.

\subsection{Opportunities: Repurposing GPUs and Migrating Requests}
\label{subsec:opportunity}

\S\ref{subsec:gpu_underutilization} shows how both idleness and decode inefficiency cause significant waste in the colocated RL system. 
We propose two complementary mechanisms, repurposing and migration, to reclaim these wasted GPU-hours.
Figure~\ref{fig:system_timeline} illustrates with data from a real run how these mechanisms complement each other.

\textbf{Repurposing} allows idle GPUs to begin training on completed GRPO groups without waiting for the entire rollout phase to finish. Since training is compute-bound, it can efficiently utilize these otherwise idle GPUs, reclaiming the idle loss it would have otherwise incurred.

\textbf{Migration} improves inference efficiency by consolidating the remaining rollouts onto fewer engines, allowing for a larger batch size and thus higher decode throughput.
Meanwhile, the GPUs that migration vacates can in turn be repurposed for training.
This optimization can reclaim the lost GPU-hours from decode inefficiency loss.

\subsection{Challenges} 
\label{subsec:challenges}
Realizing repurposing and migration in a colocated RL system raises two main challenges.

\paragraph{Challenge 1: Migration Across Diverse Workloads and Hardware}
Determining the best time to migrate an inference engine is challenging.
The ideal migration point maximizes GPU utilization in both rollout and training, but no one point works across workload and hardware configurations. 
It is difficult to determine this point analytically for various reasons:  
\begin{itemize}[leftmargin=*, nosep]
 \item \textbf{Hardware} sets KV cache capacity, memory bandwidth, and compute bandwidth.
 This directly affects the inference vs.~training throughput.
 \item Both the \textbf{Model} and \textbf{Dataset} affect the token length distribution of rollouts.
 \item \textbf{Agentic settings} add extra costs from tool-calling times, as well as extra prefill cost from appending the output of a tool call to the context.
\item \textbf{Model learning} shifts the rollout distribution from step to step as the policy learns.
\end{itemize}
The system therefore needs a \emph{low-cost online policy that picks a good migration point during the rollout phase}, rather than a static value tuned offline for one setting.

\paragraph{Challenge 2: RL Training with Dynamic GPU Allocation}
\label{subsec:challenges_repurposing}

Flexible repurposing of GPUs introduces new challenges when implementing the training phase.

First, training requires more memory than inference, and thus typically uses a different parallelism configuration.
For example, the DAPO Math $\times$ H200 configuration runs a data-parallel inference engine on each of eight GPUs, while its training phase runs four data-parallel training engines, each with TP=2.
As a result, it is not always possible to immediately repurpose inference engines since their GPUs may not fit a training engine. 
Similarly, the repurposing granularity depends on the training backend.
With FSDP, training can begin once enough GPUs are free to collectively hold the trainer state; after that, GPUs can join one at a time, at some reconfiguration cost.
With data parallelism composed with model parallelism (e.g., Megatron), training needs at least one complete model-parallel group, and capacity grows only in whole replicas.
Repurposing must therefore free this many GPUs to be worthwhile.

Meanwhile, work assignment to training engines cannot know the set of rollouts ahead of time.
In synchronous colocated execution, all samples exist before training starts, so they can be divided evenly across replicas upfront.
When GPUs are repurposed mid-rollout, completed groups instead arrive one at a time, replicas join at different times, and the longest groups arrive last.
Because every replica must finish before the shared gradient synchronization, a replica that receives a large share of these late groups delays all others; the trainer must therefore assign work online so that the tail of the rollout is balanced across replicas.

\section{\sys}
To address the challenges described in \S\ref{subsec:challenges} we introduce \sys, an on-policy RL framework that automatically maximizes GPU utilization.
Figure \ref{fig:system_overview} illustrates \sys's two primary components:
\providecommand{\stepball}{}
\DeclareRobustCommand{\stepball}[1]{\tikz[baseline=(c.base)]\node[circle, fill=black, text=white,
  inner sep=0pt, minimum size=1.15em, font=\scriptsize\bfseries] (c) {#1};}
\definecolor{boxgray}{HTML}{EFEFEF}
\definecolor{inpool}{HTML}{D4E1F2}
\definecolor{trpool}{HTML}{EBB576}
\definecolor{swL}{HTML}{A6C2E0}
\definecolor{swR}{HTML}{E8B070}
\definecolor{note}{HTML}{F9E3A1}
\definecolor{fold}{HTML}{D2BE85}
\definecolor{purp}{HTML}{8E7CC3}
\definecolor{lblue}{HTML}{AAC8F0}
\definecolor{cellg}{HTML}{999999}
\definecolor{Bred}{HTML}{C0392B}
\definecolor{barin}{HTML}{6FA8DC} \colorlet{barout}{barin!22}  

\begin{figure}[t]
\centering
\resizebox{\columnwidth}{!}{\begin{tikzpicture}[x=1cm, y=-1cm,
  frame/.style={draw=black, line width=1.4pt},
  arr/.style={-{Stealth[length=3.4mm,width=2.8mm]}, line width=2pt},
  darr/.style={arr, dashed},
  ball/.style={circle, fill=black, text=white, inner sep=0pt, minimum size=5mm, font=\large\bfseries},
  it/.style={font=\Large\itshape, inner sep=1pt},
  sit/.style={font=\large\itshape, inner sep=1pt},
  pill/.style={rounded corners=2pt, minimum width=1.45cm, minimum height=0.4cm, inner sep=0pt, line width=0.8pt},
]
\newcommand{\cells}[5]{\foreach \k in {1,...,#3} \filldraw[fill=#4, draw=black, line width=1pt] ({#1+(\k-1)*#5},{#2}) rectangle ++(#5,0.5);}
\newcommand{\strip}[5]{\foreach \k in {1,...,#3} \filldraw[fill=cellg, draw=black, line width=0.6pt] ({#1+(\k-1)*#5},{#2}) rectangle ++(#5,0.24);
  \ifnum#4>0 \foreach \k in {1,...,#4} \filldraw[fill=purp, draw=black, line width=0.6pt] ({#1+(#3+\k-1)*#5},{#2}) rectangle ++(#5,0.24);\fi}
\newcommand{\gpu}[4]{\filldraw[fill=white, draw=black!45, line width=1pt, rounded corners=3pt] (#1,#2) rectangle ++(1.75,1.5);
  \node[font=\large, text=black!65] at ({#1+0.875},{#2+0.3}) {GPU};
  \node[pill, #3] at ({#1+0.875},{#2+0.72}) {Infer};
  \node[pill, #4] at ({#1+0.875},{#2+1.17}) {Train};}
\tikzset{
  infon/.style={draw=blue!60!black, fill=blue!22, text=blue!40!black, font=\large\bfseries},
  trnon/.style={draw=orange!75!black, fill=orange!35, text=orange!35!black, font=\large\bfseries},
  sleep/.style={draw=black!35, dashed, fill=white, text=black!40, font=\large},
  infsw/.style={draw=blue!60!black, dashed, fill=blue!7, text=blue!40!black, font=\large},
  trnsw/.style={draw=orange!75!black, dashed, fill=orange!12, text=orange!35!black, font=\large}}

\node[ball, minimum size=5.5mm] at (1.12,0.98) {1};
\node[font=\LARGE, anchor=west] at (1.42,0.93) {prompts};
\filldraw[frame, fill=boxgray] (0.8,1.75) rectangle (6.95,6.77);
\node[font=\LARGE, anchor=west] at (2.3,2.15) {Rollout Router};
\filldraw[draw=black, line width=1.2pt, fill=note] (4.0,2.78) -- (6.63,2.78) -- (6.63,5.97) -- (3.47,5.97) -- (3.47,3.3) -- cycle;
\filldraw[draw=black, line width=1pt, fill=fold] (3.47,3.3) -- (4.0,2.78) -- (3.88,3.22) -- cycle;
\node[font=\Large, align=center, inner sep=0pt, anchor=north] at (5.3,2.9) {Migration\\[-2pt]Policy};
\foreach \a/\t in {3.92/4.15, 4.95/4.35, 5.46/4.66}
  \filldraw[fill=barin, draw=black, line width=1pt] (\a,5.67) rectangle ++(0.4,{\t-5.67});
\foreach \a/\t in {4.43/5.42, 5.97/5.27}
  \filldraw[fill=barout, draw=barin!60!black, line width=1pt, dash pattern=on 2.5pt off 1.5pt] (\a,5.67) rectangle ++(0.4,{\t-5.67});
\draw[Bred, line width=2pt, dash pattern=on 5pt off 4pt] (3.5,5.12) -- (6.55,5.12);
\node[text=Bred, font=\large\bfseries] at (6.17,4.7) {B};
\draw[arr] (2.05,1.33) -- (2.05,7.5);
\foreach \y in {3.25,4.1,4.95} \strip{1.72}{\y}{3}{0}{0.22}
\node[sit, anchor=west] at (2.36,6.22) {requests};
\node[ball, draw=boxgray, line width=1.2pt] at (2.05,6.22) {2};
\draw[darr] (4.97,7.55) -- (4.97,6.0);
\node[ball] at (3.45,7.15) {3};
\node[sit, anchor=east, align=right] at (4.87,7.15) {load\\[-3pt]metrics};
\filldraw[frame, fill=inpool] (0.8,7.6) rectangle (5.35,10.47);
\shade[left color=swL, right color=swR] (5.65,7.6) rectangle (9.15,10.47);
\draw[frame] (5.65,7.6) rectangle (9.15,10.47);
\filldraw[frame, fill=trpool] (9.4,7.6) rectangle (13.42,10.47);
\gpu{0.98}{7.9}{infon}{sleep}
\gpu{3.35}{7.9}{infon}{sleep}
\gpu{6.5}{7.92}{infsw}{trnsw}
\gpu{9.6}{7.95}{sleep}{trnon} \gpu{11.45}{7.95}{sleep}{trnon}
\node[it] at (3.08,9.92) {rollouts};
\node[it] at (7.45,9.92) {repurposing};
\node[font=\itshape, inner sep=1pt] at (11.41,9.95) {\tikz[baseline=(c.base)]\node[ball](c){7}; accumulate gradients};
\node[font=\LARGE] at (2.95,10.95) {Inference Pool};
\node[font=\LARGE] at (11.45,10.95) {Training Pool};

\node[rotate=90, font=\large\itshape, text=black!55] at (0.3,4.26) {host (CPU)};
\node[rotate=90, font=\large\itshape, text=black!55] at (0.3,8.45) {GPUs};
\draw[arr] (6.0,7.6) -- (6.0,6.79);
\node[ball] at (5.5,7.18) {m};
\strip{6.13}{7.06}{1}{3}{0.17}
\draw[darr] (6.63,5.47) -- (7.55,5.47) -- (7.55,7.58);
\node[ball] at (7.98,7.12) {4};
\node[sit, anchor=west] at (8.27,7.1) {repurpose};
\filldraw[frame, fill=boxgray] (8.15,1.75) rectangle (13.42,6.77);
\node[font=\LARGE, anchor=west] at (8.35,2.15) {Train Router};
\draw[arr] (6.95,4.3) -- (8.42,4.3);
\node[ball] at (7.68,3.88) {5};
\filldraw[draw=black, line width=1.2pt, fill=white!60!boxgray] (8.42,2.78) rectangle (10.72,6.45);
\node[font=\itshape, anchor=north west, inner sep=1pt] at (8.5,2.88) {buffer (GRPO)};
\foreach \y/\n in {3.75/4, 4.15/7, 4.55/2, 4.95/5} \strip{8.85}{\y}{2}{\n}{0.19}
\draw[line width=1pt] (8.75,3.72) -- (8.65,3.72) -- (8.65,5.22) -- (8.75,5.22);
\node[draw=orange!75!black, fill=orange!30, rounded corners=2pt, line width=0.8pt, font=\small, align=center, inner sep=2pt] at (9.57,5.82) {compute\\[-2pt]advantages};
\draw[arr] (10.72,4.3) -- (11.0,4.3);
\filldraw[draw=black, line width=1.2pt, fill=white!60!boxgray] (11.0,2.78) rectangle (13.2,6.45);
\node[font=\itshape, anchor=north west, inner sep=1pt] at (11.08,2.88) {batches};
\foreach \y/\n in {3.45/4, 4.85/2, 5.7/1} {
  \foreach \r in {1,...,\n} \strip{11.5}{\y+0.1+(\r-1)*0.27}{2}{4}{0.2}
  \draw[orange!75!black, line width=1.2pt, rounded corners=2pt] (11.38,\y) rectangle (12.82,{\y+0.18+\n*0.27});
}
\draw[arr] (12.1,6.45) -- (12.1,7.58);
\node[ball] at (12.45,7.12) {6};
\node[sit, anchor=west] at (12.73,7.1) {batches};
\draw[arr] (13.42,9.45) -- (13.8,9.45) -- (13.8,11.75) -- (0.42,11.75) -- (0.42,9.45) -- (0.78,9.45);
\node[fill=white, inner sep=3pt, font=\large] at (7.1,11.75)
  {\tikz[baseline=(c.base)]\node[ball](c){8}; \textit{optimizer step $\rightarrow$ weight sync}};
\end{tikzpicture}}
\caption{\sys{} overview mid-step. Each GPU hosts an inference and a training engine
  (one live, one asleep); routers run on the host.
  \stepball{1}~Prompts arrive at the Rollout Router, which \stepball{2}~dispatches them
  to inference GPUs that \stepball{3}~report load. When a GPU's live rollouts fall below
  $B$ decided by the migration policy (\S\ref{s:migration_policy}), its requests are \stepball{m}~re-dispatched and the
  GPU is \stepball{4}~repurposed for training. \stepball{5}~Completed rollouts stream to
  the Train Router (\S\ref{s:training_on_streaming_inputs}), which computes GRPO
  advantages and \stepball{6}~dispatches batches. \stepball{7}~Gradients accumulate;
  \stepball{8}~after the last rollout, the optimizer steps and weights sync via host memory.}
\label{fig:system_overview}
\end{figure}
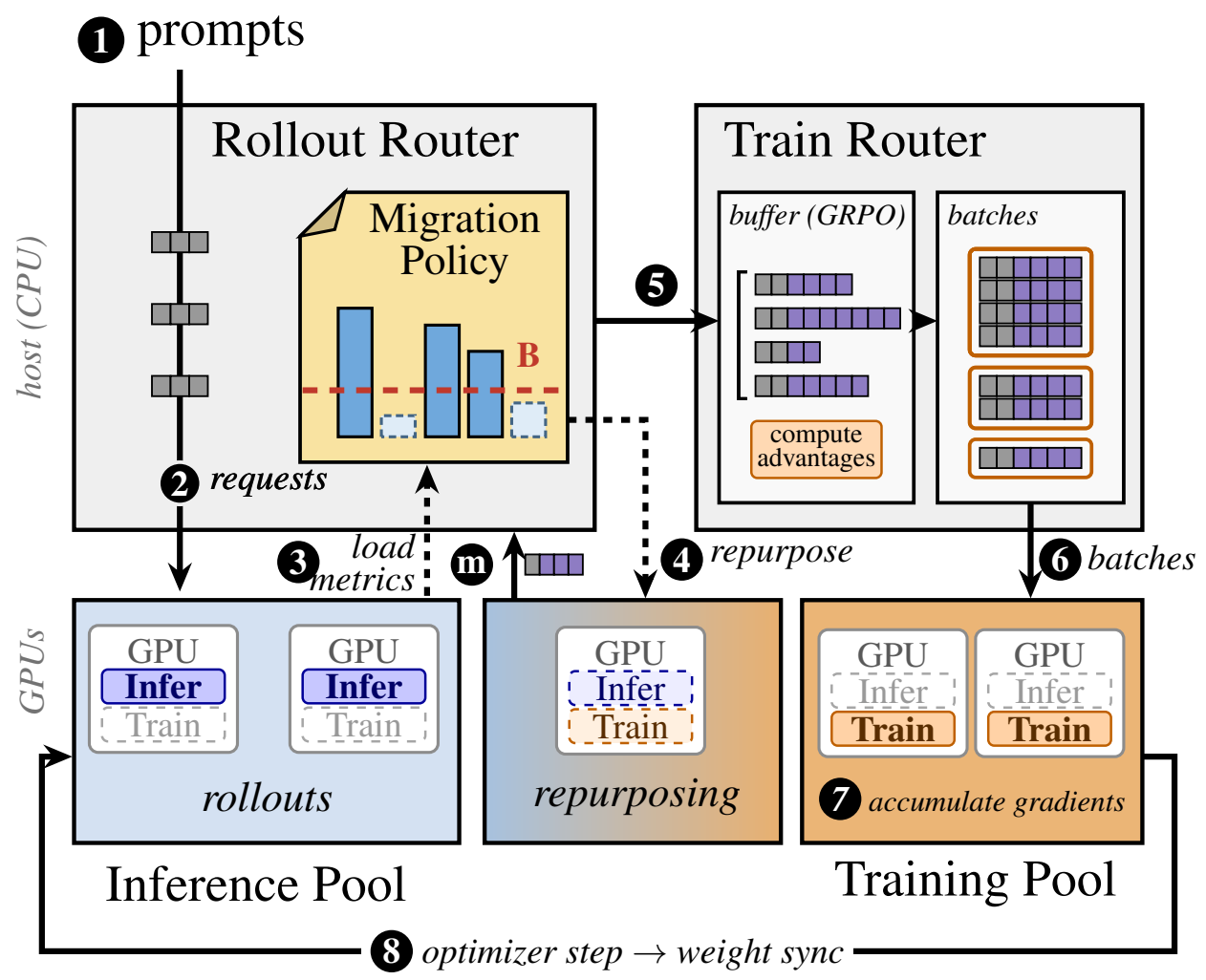 \begin{itemize}[noitemsep, topsep=0pt]
    \item The \emph{Rollout Router} orchestrates the inference engines. It assigns rollout work to inference engines, uses a \emph{migration policy}~(\S\ref{s:migration_policy}) to decide when to migrate work away from an engine, and repurposes inference engines for training by handing them to the \TR{}.
\item The \emph{Train Router} orchestrates the training engines. It dynamically incorporates new GPUs into the trainer, assigns training work to replicas, and synchronizes the training replicas~(\S\ref{s:training_on_streaming_inputs}).
\end{itemize}

At the beginning of each RL step, the \RR{} distributes the batch of prompts across inference engines.
As inference engines generate rollouts, they continuously report load metrics to the \RR{}.
The \RR{} consults the migration policy to detect when to trigger migration and repurposing.
When an inference engine meets migration conditions, the \RR{} preempts the engine's running requests, migrates them to the least-loaded engine, scales down the now idle inference engine, and hands ownership of the GPUs to the \TR{}.

The \TR{} consumes rollouts produced by the \RR{} and dispatches batches to training engines for gradient generation.
Once all rollouts in the batch have been processed by training, \sys performs the configured optimizer step and weight synchronization.
The process repeats for the next step.

In the rest of this section, we describe \sys's key design features: 
a migration policy that adaptively finds the best migration point for a dynamic workload while remaining training-aware (\S\ref{s:migration_policy}), and a parallelism-aware \TR{}~(\S\ref{s:training_on_streaming_inputs}).

\subsection{Migration Policy}
\label{s:migration_policy}

The simplest policy for repurposing is to disable migration entirely and repurpose each inference engine when it finishes its rollouts.
However, this wastes efficiency from under-utilized inference GPUs.
\sys's migration policy thus aims to identify these under-utilized GPUs and compact their work to maximize overall utilization.

Determining the conditions for migration is therefore key to maximizing GPU utilization.
We frame this problem as a search for a threshold $B$: when the number of remaining rollouts falls below $B$ on an inference engine, \sys triggers a migration from that engine and repurposes the victim engine's GPUs for training.

\begin{figure}[t]
    \centering
    \includegraphics[width=\columnwidth]{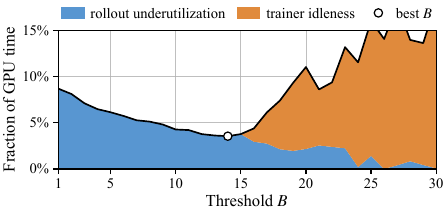}
    \caption{
    Simulated (\S\ref{s:eval:policysim}) performance loss from the utilization-optimal at different values of $B$ for a single step of DAPO-Math. 
    Migrating too early (higher $B$) and migrating too late (lower $B$) both result in underutilization.
    }
\label{fig:migration_policy_landscape}
\end{figure}

As Figure~\ref{fig:migration_policy_landscape} illustrates, 
late migrations (low values of $B$) cause inference engines to linger at low utilization.
Meanwhile, early migrations (high values of $B$) can reduce inference throughput prematurely and create a bottleneck for training, again leading to underutilization.
As a result, the optimal migration point is somewhere in the middle.

\subsubsection{Finding $B$}
\label{subsec:finding_b}
The optimal $B$ depends on workload and hardware characteristics.
To discover the optimal $B$ value, \sys uses a simple linear controller inspired by delay-based network congestion controllers~\cite{vegas}, shown in Algorithm \ref{alg:batch_migration} alongside the associated migration logic.
Intuitively, the controller should set $B$ such that inference vs.~training throughput are evenly matched, i.e.~neither phase is the bottleneck.
Training is compute-bound and its inputs are known ahead of time, so training throughput is relatively easy to estimate.
On the other hand, rollouts generate inference work dynamically and it is challenging to estimate the maximum possible inference throughput for a given rollout a priori, as this is workload-dependent~(\Cref{subsec:challenges}).
Thus, the main challenge is finding a reliable signal for the balance point between inference and training.

We observe that increasing $B$ generally improves total GPU utilization until the trainer begins to starve~(\Cref{fig:migration_policy_landscape}).
Thus, the best $B$ is the largest value that avoids training starvation.
Meanwhile, training starvation is also easy and robust to measure online; we simply check for idle time on each training GPU.
Therefore, \sys uses starvation as a binary feedback signal.
After each RL step, if \emph{any} training starvation occurred, the controller decreases $B$ by a configurable fixed step $\Delta B$; otherwise, it increases $B$ by $\Delta B$.
Because the controller updates $B$ every step, it requires no offline profiling or user tuning and adapts as the rollout distribution shifts over the course of training.

\sys must also pick $B_0$, the initial value of $B$.
A single value may not work across all workloads and hardware environments.
Two reasonable defaults would be the minimum and maximum possible $B$.
To this end, we observe that the two extremes Figure~\ref{fig:migration_policy_landscape} illustrates are not equally costly;
while inference engines still make progress with low batch sizes, starved training engines make no progress at all.
As a result, a high-$B$ policy (migrate too early) would be worse than a low-$B$ policy (migrate too late).
So, \sys conservatively sets $B_0$ to the controller's step size by default, although $B_0$ can be manually configured if desired.
As we will show in \S\ref{s:eval:policysim}, even with a conservative $B_0$ \sys delivers good performance over a range of workloads and hardware configurations.

\subsubsection{KV cache contention}
\label{subsec:kv_cache_contention}

While our evaluation shows that training starvation is a reliable signal for balancing rollout vs.~training throughput, one shortcoming is that we can only apply the signal to the \emph{subsequent} rollout steps.
Thus, if a particular step deviates significantly in its optimal $B$ from the previous step---for example, due to an unexpected straggler---then \sys may migrate earlier than is optimal.
Also, \sys cannot read the training starvation signal until trainers begin execution; so, if migration happens too early, the described policy cannot recover for the current step.
This manifests in slowdowns from overloaded inference engines.

As a safeguard for such cases, we supplement the inter-step controller with intra-step information that we can collect at the intended migration time.
In particular, we 
add the \emph{KV cache veto} (line 5 of Algorithm \ref{alg:batch_migration})
which  
estimates whether a migration destination is at capacity by reading the receiving engine's KV cache capacity
and delays rollout migration (line 6) if it is saturated.
By the end of the step, this procedure 
reduces $B$ to the maximum value that would avoid saturating the most constrained KV cache.
This helps constrain the range of $B$ to values which trigger migrations with ample KV cache capacity.

\begin{algorithm}[tb]
\caption{Batch-threshold repurposing and linear control algorithm}
\label{alg:batch_migration}
\begin{algorithmic}[1]

\STATE \textit{// On rollout completion}
\FOR{each engine $e \in RolloutEngines$}
  \IF{$\textsc{NumRollouts}(e) <= B$}
    \STATE $dest \gets \textsc{LeastLoaded}(RolloutEngines)$
    \IF{$\textsc{KV}(e) + \textsc{KV}(dest) > 100\%$}
        \STATE $B$ -= $\Delta B$
        \STATE \textbf{continue}
    \ENDIF
    \STATE \textsc{Migrate}$(dest, e)$
\STATE \textsc{RepurposeToTraining}$(e)$
  \ENDIF
\ENDFOR
\STATE
\STATE \textit{// On full step completion}
\IF{$\textsc{StarvationDetected}(TrainEngines)$}
  \STATE $B$ -= $\Delta B$
\ELSE
  \STATE $B$ += $\Delta B$
\ENDIF

\end{algorithmic}
\end{algorithm}
 
\subsubsection{Training-aware migration}
\label{subsec:migration_constraints}
Training in \sys remains on-policy 
as long as there is barrier between rollout generation and the optimizer step.
So, to ensure on-policy training, \sys defers the optimizer step and all nonlinear whole-batch operations such as gradient clipping until all gradients have been accumulated.

Training engine configurations can 
both introduce reconfiguration overheads and
affect 
whether a migration results in a speedup.
For example, a Megatron training engine with TP=2 cannot incorporate a single GPU to do training, meaning vacating only one GPU does not allow any repurposing to happen.
Likewise, an FSDP training engine may not be able to fit unsharded model weights on a single GPU.
To avoid wasteful migrations, the \RR{} only triggers a migration if there exists a set of migration-eligible inference engines which can immediately contribute to training.

Similarly, the \RR{} must consider the overhead of adding GPUs to training.
FSDP shards model weights across all GPUs, so adding a GPU to an FSDP training engine requires pausing to redistribute weights.
To reduce the number of training reconfigurations, the \RR{} can bundle GPU migrations which happen in quick succession into a single event so the \TR{} can add all GPUs with a single reconfiguration.

\subsection{Train Router}
\label{s:training_on_streaming_inputs}

Synchronous RL systems know the entire training workload when training starts, allowing for optimal work distribution between training engines.
With \sys, training engines must process a stream of rollouts on a dynamically changing group of training GPUs.
Because rollout lengths can vary, it is possible for streaming rollouts to cause workload imbalance on training engines.
For example, longer rollouts tend to finish later.
Thus, if independent data-parallel trainers pull batches of work greedily from the work queue with no load-balancing mechanism, a small subset of training engines then train on a batch of much longer requests. 
This causes starvation among other training workers, which harms performance (Table~\ref{tab:ablation_DAPO_h200}).

To address this challenge, \sys includes a \TR{}, which ensures that 
data-parallel training engines pull gradually smaller batches near the end of an RL step. 
At the beginning of a step, the \TR{} sends $64$ rollouts at a time to training engines.
As the number of remaining rollouts decreases, the \TR{} sends at most one fourth of the remaining rollouts to the training engines.
Reducing the training batch size given to engines only towards the end of training outperforms both 
the naive work-distribution scheme that causes trainer idling 
and
an alternative policy that uses small training batches throughout training. This is because the \TR{} must do some fixed work per batch, so always using small batches incurs overhead.

 \section{Implementation}
\label{s:impl}
We support two training backends to show \sys's applicability across workloads.
\sysm builds on the Slime~\cite{Slime} post-training framework and uses SGLang for inference and Megatron~\cite{Megatron} for training.
\sysf, specialized for memory-constrained hardware, builds on the SkyRL~\cite{skyrl} post-training framework and uses vLLM for inference and FSDP2~\cite{fsdp} for training.
\sysf additionally applies memory-saving optimizations to SkyRL to minimize the memory footprint of training.
Most significantly, it completely offloads the optimizer to CPU memory akin to ZeRO-Offload~\cite{zerooffload}, allowing training to begin even when limited GPU memory is available.
\sysf also enables sequence packing of variable-length sequences~\cite{packing}, allowing denser microbatches.
Both \sysm and \sysf use Ray~\cite{Ray} actors to encapsulate GPU workers, use SkyRL-Gym for tool-calling support, and use the Adam optimizer~\cite{AdaM}.

\paragraph{Rollout Router}
The Rollout Router assigns rollouts to inference engines and manages request migration once \sys repurposes inference engines for training.
It implements group affinity, which ensures the same inference engine handles all of a GRPO group's samples.
This increases prefix cache hit rates, which are especially important for multi-turn trajectories.

\paragraph{Request Migration}
To trigger a migration, the \RR{}
\begin{inparaenum}[(i)]
\item preempts all in-flight rollouts,
\item captures any generated tokens, and
\item sends them to the least-loaded remaining inference engine.
\end{inparaenum}
We experimented with using KV-cache occupancy to select migration destinations--including distributing requests across multiple engines--but found that the least-loaded inference engine policy performed best.
We choose to migrate tokens rather than KV cache contents since it is simpler and offered the same performance.

\paragraph{Repurposing GPUs}
While GPUs process rollouts, \sys swaps all training state to CPU memory using the base RL system's existing mechanisms.
\sysm uses data parallelism (DP) and tensor parallelism (TP), so TP groups are independent and synchronization across groups is not needed.
In \sysf, adding a new FSDP rank requires re-sharding training state and re-building NCCL communicators.
So, when \sysf triggers a reconfiguration, it waits for existing workers to complete their assigned microbatches, then gathers the sharded gradient in a pinned CPU buffer.
Then, it assigns new sharding keys to the new set of train workers, which then pull their corresponding gradient and weight shards from CPU memory.
This adds 2-3\% overhead overall to \sysf. Future work might reduce this overhead with direct GPU-GPU communication.
 \section{Evaluation}
\label{sec:evaluation}

To evaluate \sys, we ask the following questions:
\begin{enumerate}[noitemsep, topsep=0pt]
\item How do the end-to-end speedups \sys provides compare to prior RL training systems? (\S\ref{s:eval:e2e})
\item How do \sys's components contribute to the speedups it offers? (\S\ref{s:eval:ablations})
\item Does \sys's controller discover effective migration thresholds, despite workload variation? (\S\ref{s:eval:policysim})
\end{enumerate}

\subsection{Setup}
\paragraph{Applications}
\label{s:eval:setup:apps}
We evaluate \sys on two applications: 
SkyRL's Text2SQL dataset~\cite{skyrlsql} and DAPO-Math~\cite{DAPO}.
For Text2SQL, we follow the training recipe from the SkyRL-SQL paper, fine-tuning Qwen2.5-Coder-7B-Instruct on Text2SQL over 35 steps.
A Text2SQL rollout consists of 1-5 sequential LLM generations of at most $3,000$ tokens, interspersed with SQL requests against a SQLite database in the style of a ReAct agent~\cite{react}. We show the rollout distribution in Figure~\ref{fig:rollout_tail_distribution}(c). 

For the DAPO-Math task, we fine-tune DeepSeek-R1-Distill-Llama-8B over 20 steps.
A DAPO-Math rollout consists of a single long LLM generation of up to $32,000$ tokens.
Synchronous runs of DAPO-Math are bottlenecked by the long-tail, but to a lesser extent than Text2SQL: as shown in Figure~\ref{fig:rollout_tail_distribution}(a), the maximum observed rollout generation length is $28,000$ tokens, and the median is $6,000$ tokens.
A synchronous RL system exhibits significantly less idleness on DAPO-Math than on Text2SQL~(Table~\ref{tab:optimality_table}).
\paragraph{Hardware}
\label{s:eval:setup:hardware}
We evaluate on the following single-node testbeds: \texttt{8$\times$H200} (141\,GB HBM3e each; NVLink; 2× Intel Xeon Platinum 8480C (112 cores); 2 TB DDR5 host memory),
\texttt{4$\times$H100} SXM GPUs (80\,GB HBM3) in an NVIDIA DGX H100 node, connected all-to-all by fourth-generation NVLink through NVSwitch (900\,GB/s per GPU), with 2$\times$ Intel Xeon Platinum 8480C CPUs (112 cores) and 2\,TB of host memory. \texttt{4$\times$L40S GPUs} (48\,GB GDDR6 each) connected over PCIe Gen4 (no NVLink), with 2$\times$ AMD EPYC 9954 CPUs (128 cores) and 1.5\,TB of host memory. We refer to each testbed by its GPU configuration.
\paragraph{System Configurations}
We cover both memory-abundant and memory-constrained settings.
Memory-abundant settings use Megatron with TP=2 and DP on the H200 testbed.
Memory-constrained settings use FSDP, running Text2SQL and DAPO-Math on the L40S and H100 testbeds respectively.
Unless stated otherwise, \sys's migration policy uses $B_0=8$ and $\Delta B=8$~(\Cref{s:migration_policy}).

\subsection{End to End Results}
\label{s:eval:e2e}

\begin{figure}[t]
\centering
\includegraphics[width=\columnwidth]{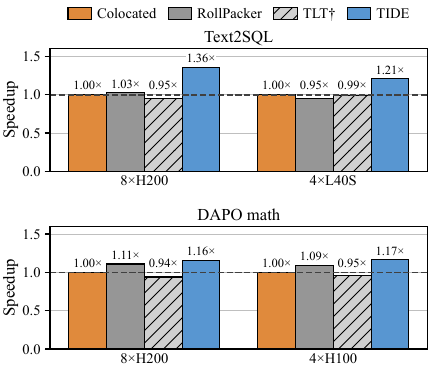}
\caption{End-to-end speedup over fully synchronous colocated execution (baseline step time divided by system step time).
  $^\dagger$TLT's speedup is relative to its own run without speculative decoding.}
\label{fig:e2e_speedups}
\end{figure}

We show the results of the Text2SQL (35 steps) and DAPO Math (20 steps) experiments across the testbeds in Figure~\ref{fig:e2e_speedups}.

We compare \sys against three baselines: 
\begin{enumerate}[noitemsep, topsep=0pt, wide]
\item Synchronous colocated RL training (``Colocated'').
\item RollPacker's algorithm~\cite{RollPacker} for GPU repurposing ported onto \sys{} for fair comparison. \item TLT~\cite{hu2026tlt}, a synchronous RL system that accelerates long-tail rollouts using speculative decoding.
\end{enumerate}

To evaluate the best possible speedup TLT could offer, we made two changes. First, TLT calls for online training of the draft model, which can add run-time overhead; we instead trained TLT's EAGLE-2 draft model on application rollouts offline, eliminating any possible run-time overhead.
Second, TLT uses verl~\cite{verl} as its base RL system, which introduces substantial overheads over \sys{}'s base systems Slime and SkyRL; we thus report TLT's speedup over itself without speculation instead.

Over all workloads and testbeds, \sys{} outperforms all baselines.
On the H200 testbed it improves upon colocated execution by 1.36$\times$ on Text2SQL and 1.16$\times$ on DAPO-Math.
On the memory-constrained testbeds the speedups are 1.21$\times$ (Text2SQL) and 1.17$\times$ (DAPO Math).
We also see that, on all workloads, \sys{}  recovers a large amount of the optimal throughput, ranging from 77.2\% to 97\% (\Cref{tab:e2e_of_optimal}).
Text2SQL broadly exhibits larger speedups than DAPO-Math, as multi-turn trajectories increase the dispersion of rollout lengths, leading to added idle time.
Under colocated execution, $30.0\%$ of the step's GPU-time is idle on Text2SQL against $15.2\%$ on DAPO Math (\S\ref{subsec:gpu_underutilization}), so there is twice as much idle time for \sys{} to reclaim.

\begin{table}[t]
\centering
\caption{Fraction of utilization-optimal throughput (optimal / system step time).}
\label{tab:e2e_of_optimal}
\begin{tabular*}{\columnwidth}{@{\extracolsep{\fill}}llcc@{}}
  \toprule
  Application               & GPUs          & Colocated & \sys{}          \\
  \midrule
  DAPO Math                 & 8$\times$H200 & 75.1\%    & \textbf{87.0\%} \\
  \quad R1-Distill-8B       & 4$\times$H100 & 83.2\%    & \textbf{97.4\%} \\
  \addlinespace[2pt]
  Text2SQL                  & 8$\times$H200 & 56.6\%    & \textbf{77.2\%} \\
  \quad Qwen2.5-Coder-7B    & 4$\times$L40S & 78.4\%    & \textbf{95.1\%} \\
  \bottomrule
\end{tabular*}
\end{table} 
\textbf{Rollpacker} harvests some idle GPU time, but due to its one-shot coarse migration of exactly half of all GPUs, it regularly leaves training or rollout generation underutilized.

\textbf{TLT}'s speculative decoding (SD) yields poor end-to-end performance, as improvements in tail decoding speed are heavily workload-dependent and offset by TLT's overheads.
The draft model achieves mean acceptance lengths of 4.86 and 2.25 (out of a maximum of 9) on Text2SQL and DAPO-Math, against break-even length of 2.48 (4$\times$L40s) and 2.25 (4$\times$H100), respectively: speculation accelerates tail decoding by 64\% on Text2SQL but only breaks even on DAPO-Math.
However, because SD can reduce inference throughput when the workload is compute-bound, TLT only enables SD below a certain inference batch size.
Unfortunately, this phase switch between SD enabled vs.~disabled still incurs overheads including an additional prefill with the draft model at the phase switch and the inability to selectively enable certain engine optimizations only when SD is disabled.
Finally, the draft model adds memory overhead, which can slow down training in memory-constrained settings.
Thus, in all evaluated cases, TLT's performance actually regresses from that of its colocated counterpart.

\begin{figure}[t]
\centering
\includegraphics[width=\linewidth]{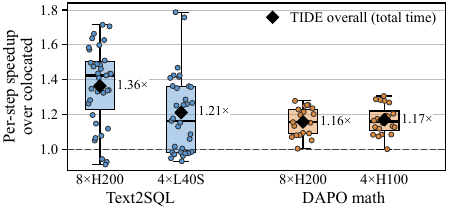}
\caption{Per-step speedup of \sys{} over colocated (points: steps; diamonds:
  overall; dashed: 1$\times$). The median step is faster in every
  configuration, with overall speedups of 1.16--1.36$\times$.
}
\label{fig:speedup_boxplot}
\end{figure}

\paragraph{Step-by-step speedups}
Digging deeper into \sys's performance, we show how \sys's gains vary across RL steps.
Figure~\ref{fig:speedup_boxplot} shows the distribution of per-step speedups for all 4 workloads.
On the H200 testbed, \sys achieves speedups in 100\% of DAPO Math steps and 91\% of Text2SQL steps.
The speedups in Text2SQL are more variable
because the Text2SQL rollout distribution is more skewed~(\Cref{fig:rollout_tail_distribution}); in some steps, a few straggler rollouts allow \sys to almost entirely hide training work.

The memory-constrained settings (4xL40S and 4xH100) use \sysf to fit training on the available resources.
Because repurposing for FSDP incurs more reconfiguration overheads than repurposing for DP, we sometimes see neutral or negative speedup on Text2SQL on the L40S testbed, in particular when there are fewer stragglers.
Fixed repurposing costs are more exposed in Text2SQL, which has shorter steps than DAPO.
In aggregate, \sysf still achieves a significant speedup over the colocated baseline.

\subsection{Ablation Study}
\label{s:eval:ablations}
To measure the contribution of each \sys feature, we enable them one at a time and evaluate the speedup on DAPO-Math (Table~\ref{tab:ablation_DAPO_h200}), on both the 8$\times$H200 testbed and the memory-constrained 4$\times$H100 testbed.
The features are, in order:
\emph{repurposing}, which enables inference engines to release their GPUs to training as soon as their rollouts finish;
\emph{load balancing}, which evenly distributes work across the training GPUs (\S\ref{s:training_on_streaming_inputs});
\emph{migration}, which migrates the remaining rollouts off underutilized GPUs once their count falls below an adaptively tuned threshold (\S\ref{s:migration_policy});
and the \emph{KV veto}, which blocks migrations that would overflow the KV cache of the receiving engines (\S\ref{subsec:kv_cache_contention}).

\begin{table}[t]
\centering
\caption{Ablation on DAPO Math: seconds per step, with speedup over colocated in parentheses. Each row adds one mechanism to the row above; bold marks the best per column.}
\label{tab:ablation_DAPO_h200}
\vskip 0.1in
\small
\begin{tabular*}{\columnwidth}{@{\extracolsep{\fill}}lcc@{}}
  \toprule
                   & 8$\times$H200                 & 4$\times$H100                  \\
  System           & (\sysm)                       & (\sysf)                        \\
  \midrule
  Colocated        & 689.1 (1.00$\times$)          & 1142.6 (1.00$\times$)          \\
  + Repurposing    & 666.8 (1.03$\times$)          & 1025.5 (1.11$\times$)          \\
  + Load balancing & 617.9 (1.12$\times$)          & N/A$^\dagger$                  \\
+ Migration      & \textbf{572.5 (1.20$\times$)} & 1052.4 (1.09$\times$)          \\
  + KV veto        & 576.9 (1.19$\times$)          & \textbf{1002.9 (1.14$\times$)} \\
  \bottomrule
\end{tabular*}

\vspace{3pt}
{\footnotesize\raggedright $^\dagger$\,\sysf{} always dispatches microbatches to FSDP workers
  with centralized bin-packing, so repurposing already includes load balancing.\par}
\end{table}

We find that repurposing GPUs alone does not yield benefits on 
the 8$\times$H200 testbed (1.03$\times$) since large batches cause load imbalance near the end of each step.
With load balancing, the speedup on 8$\times$H200 is 1.12$\times$, and on the 4$\times$H100 testbed the speedup is 1.11$\times$.

Adding migration raises the 8$\times$H200 speedup from 1.12$\times$ to 1.20$\times$: consolidating the remaining rollouts onto fewer engines frees GPUs for training earlier, instead of waiting for each engine to drain on its own.

Finally, while the KV veto has no measurable effect in the 8$\times$H200 testbed since inference engines are well-provisioned and rarely experience KV-cache pressure,
it is essential 
on the memory-constrained 4$\times$H100 testbed.
Without it, migration is \emph{slower} than load balancing alone ($1.09\times$ vs. $1.11\times$)
because 
when KV caches are full, migrations add marginal benefits but incur overheads from contention.
Adding the KV veto prevents these unnecessary migrations, increasing the overall speedup to 1.14$\times$.\footnote{
Speedups differ from end-to-end results as the ablation sampled a fresh set of rollouts with different lengths.
}

\subsection{Policy Study}
\label{s:eval:policysim}
\begin{figure}[t]
\centering
\includegraphics[width=\linewidth]{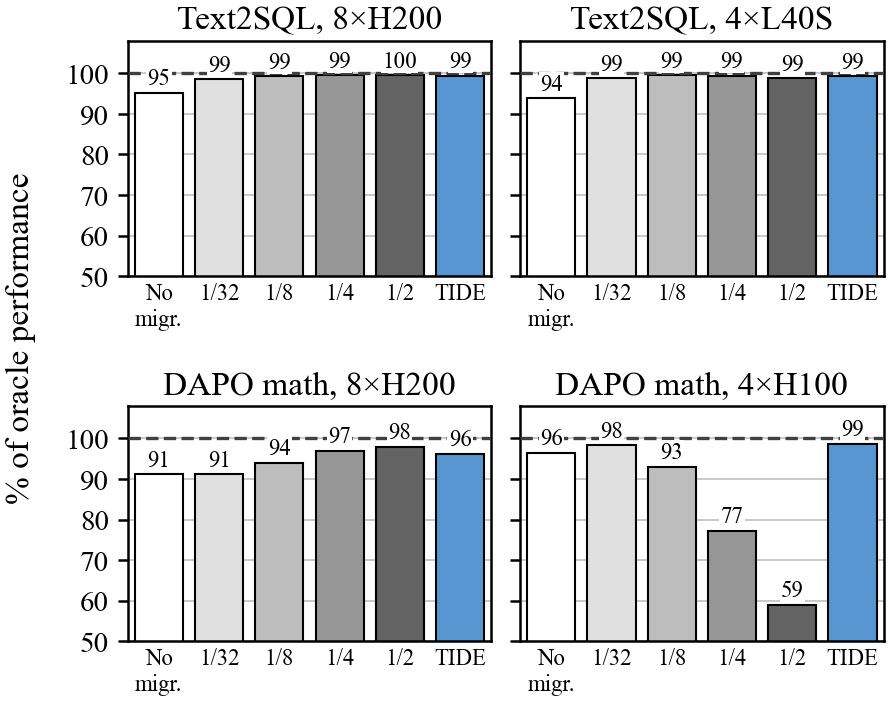}
\caption{Simulated performance of fixed-$B$ policies and \sys's adaptive policy, relative to an oracle that picks the best $B$ every step (higher is better). Fixed policies set $B$ to the given fraction of the initial per-GPU batch size. The best fixed $B$ differs across workloads, and a poor choice costs much more on some (DAPO Math) than others (Text2SQL); \sys{} is comparable to the best fixed policy without manual tuning.
}
\label{fig:policy_versus_fixed}
\end{figure}
 We study how effective \sys's controller is in finding the optimal migration threshold.
To quickly search a wide range of policies, we use ServingStudio~\cite{servingstudio} to simulate different migration policies.
In simulation, we replay a collected trace using every possible $B$ value~(\Cref{s:migration_policy}).
We define the oracle policy as one that uses the best $B$ value at each step, and compare various fixed policies and \sys's controller to this oracle.

Figure~\ref{fig:policy_versus_fixed} shows the results.
While at least one fixed policy can achieve most of the oracle's performance, the best fixed policy varies by workload.
\sys's adaptive policy provides performance comparable to that of the best fixed policy across all the workloads we evaluate without manual configuration.
 \section{Future Work}
\sys was evaluated on single node configurations with varying capability levels of hardware. 
Meanwhile, hardware innovation continues to move at break-neck speed,
from NVIDIA's NVL72 clusters, in which much larger amounts of GPUs are connected via high-speed NVLink interconnects \cite{NVL72_GB200, NVL72_GB300}, to wafer-scale chips like Cerebras' CS-4 systems \cite{Cerebras}. 
The hardware that is used changes key factors, such as the ratio of inference to training throughput, which affects the optimal migration time.
Thus, adaptive policies such as those introduced by \sys{} are critical for adapting to new hardware.
Additionally, when operating at scale, stragglers have even more impact, offering larger opportunities for \sys to reclaim GPU-hours.

\sys is also complementary to other techniques that accelerate the long tail in RL, such as speculative decoding.
While \sys migration enlarges batch sizes, speculative decoding can realize further gains, especially when migration is no longer useful and at small batch sizes.

The ideas in \sys extend beyond on-policy, colocated deployments.
They could be used to improve the performance of disaggregated deployments where the number of GPUs in the inference and training pools is traditionally static, such as a one-step overlap async RL policy; in this deployment, even with the overlapping of training and inference, the optimal split of inference and training pools changes over time in order to keep the GPUs fully utilized. 
The flexible allocation in \sys can translate to help manage this split dynamically while preserving the semantics.
In other deployments, such as fully-async RL, the GPUs are able to be kept at much higher utiilization in both the inference and training pools; thus, while there might be some benefits from flexibly allocating inference and training pools, we expect there to be less systems-level benefit from \sys.

 \section{Conclusion}

In this paper, we identified and quantified two main sources of GPU underutilization in colocated on-policy training: idle GPUs waiting for straggler rollouts (\textit{idle loss}) and inefficient decoding from low batch sizes (\textit{decode inefficiency loss}). We presented \sys, a system that improves GPU utilization in synchronous colocated RL while preserving on-policy training semantics. \sys overlaps training with ongoing rollouts by repurposing GPUs and consolidates remaining rollouts through request migration. An adaptive migration policy balances inference and training without workload-specific tuning. Across diverse workloads, hardware configurations, and training backends, \sys achieves up to 1.36$\times$ end-to-end speedup and up to 97\% of utilization-optimal throughput. 
\clearpage

\appendix
\section{Case Study}
\begin{figure}[!t]
\centering
\includegraphics[width=\columnwidth]{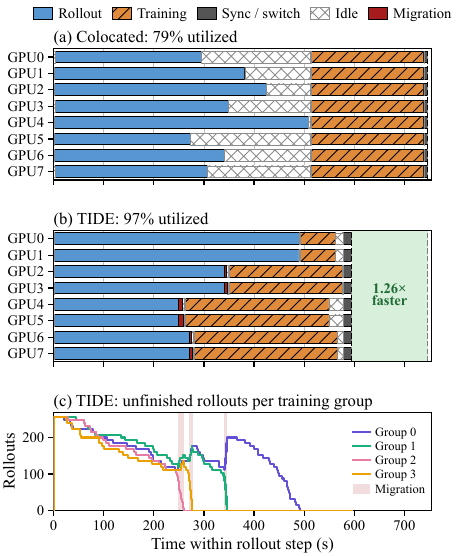}
\caption{Per-GPU timeline of a sampled rollout step under the colocated baseline (top) and \sys{} (bottom). As batch sizes shrink during the rollout tail, \sys{} consolidates in-flight requests onto fewer engines and switches the freed GPUs to training. This removes most of the idle time that colocation leaves and shortens the step.}
\label{fig:rollout_timeline_dapo}
\end{figure}

To examine the behavior of our system under a real workload, we take DAPO Math run on the 8xH200 setting as a case study.
Figure~\ref{fig:rollout_timeline_dapo} shows one step of that run on every GPU under colocated and \sys.
Each inference engine requires 1 GPU (resulting in 8 inference engines), while each training engine uses 2 GPUs (TP=2, resulting in 4 training engine data-parallel groups).
We see how, unlike the colocated, \sys is able to fill in the gaps that rollout idle time with training.
Under colocated execution, $20.8\%$ of the step's GPU-time, is spent idle.
In contrast, \sys fires three migrations progressively as batch sizes decrease across the inference engines(at 249, 271 and 341\,s, moving 80, 80 and 88 requests), while flipping the now-idle GPUs to training. 
As a result, idle step time falls to $3.2\%$ of GPU-time. 
Additionally, we see that most of the remaining idle GPU hours is the result of training ranks waiting for the stragglers trainers to finish on their samples.

We also see that the decode throughput increased as a result of the migration.
Per active engine, decode throughput rose from 2{,}766 to 2{,}897\,tok/s, because after the migrations the tail is decoded at a larger batch on two engines rather than at a batch of a few requests on eight.
Counting all GPU-time spent in inference mode, the rollout costs 1.14 GPU-hours under colocated execution but 0.80 GPU-hours under \sys{}.
With these two improvements, our system allows the step to complete in 595\,s as opposed to 746\,s.

These trends generalize to the entire run as well; over the 20 steps, idle time drops from $15.2\%$ to $4.8\%$ of GPU-time and active decoding from 15.48 to 14.82 GPU-hours ($-4.3\%$).

\section*{AI Usage Disclosure}
The authors used coding agents (Claude Code) to implement parts of the system described, and to perform data analysis on exploratory features.
The authors also used LLMs (Gemini, Claude) to perform literature search.
The authors used coding agents to generate figures and evaluation code.
The authors verified the correctness of the outputs and code. 
Additionally, the authors used large language models to suggest copy editing and formatting suggestions to the paper text. 


\begin{thebibliography}{34}
\providecommand{\natexlab}[1]{#1}
\providecommand{\url}[1]{\texttt{#1}}
\expandafter\ifx\csname urlstyle\endcsname\relax
  \providecommand{\doi}[1]{doi: #1}\else
  \providecommand{\doi}{doi: \begingroup \urlstyle{rm}\Url}\fi

\bibitem[Brakmo et~al.(1994)Brakmo, O'Malley, and Peterson]{vegas}
Brakmo, L.~S., O'Malley, S.~W., and Peterson, L.~L.
\newblock {TCP Vegas: New Techniques for Congestion Detection and Avoidance}.
\newblock In \emph{Proceedings of the Conference on Communications
  Architectures, Protocols and Applications}, SIGCOMM '94, pp.\  24–35, New
  York, NY, USA, 1994. Association for Computing Machinery.
\newblock ISBN 0897916824.
\newblock \doi{10.1145/190314.190317}.
\newblock URL \url{https://doi.org/10.1145/190314.190317}.

\bibitem[Cao et~al.(2025)Cao, Li, Zhao, Yuan, Hegde, Chen, Ruan, Griggs, Liu,
  Tang, Liaw, Moritz, Zaharia, Gonzalez, and Stoica]{skyrl}
Cao, S., Li, D., Zhao, F., Yuan, S., Hegde, S.~R., Chen, C., Ruan, C., Griggs,
  T., Liu, S., Tang, E., Liaw, R., Moritz, P., Zaharia, M., Gonzalez, J.~E.,
  and Stoica, I.
\newblock {SkyRL-Agent: Efficient RL Training for Multi-Turn LLM Agent}, 2025.
\newblock URL \url{https://arxiv.org/abs/2511.16108}.

\bibitem[Chen et~al.(2026)Chen, Liu, Sun, Li, Wang, Liu, Wen, Feng, and
  Zhang]{chen2026respec}
Chen, Q., Liu, Z., Sun, P., Li, S., Wang, G., Liu, Z., Wen, Y., Feng, S., and
  Zhang, T.
\newblock {ReSpec: Towards Optimizing Speculative Decoding in Reinforcement
  Learning Systems}.
\newblock In \emph{Proceedings of the 9th Annual Conference on Machine Learning
  and Systems}, MLSys '26, 2026.
\newblock \doi{10.48550/arXiv.2510.26475}.
\newblock arXiv:2510.26475.

\bibitem[Fu et~al.(2025)Fu, Gao, Shen, Zhu, Mei, He, Xu, Wei, Mei, Wang, Yang,
  Yuan, and Wu]{AReaL}
Fu, W., Gao, J., Shen, X., Zhu, C., Mei, Z., He, C., Xu, S., Wei, G., Mei, J.,
  Wang, J., Yang, T., Yuan, B., and Wu, Y.
\newblock {AReaL: A Large-Scale Asynchronous Reinforcement Learning System for
  Language Reasoning}, 2025.
\newblock URL \url{https://arxiv.org/abs/2505.24298}.

\bibitem[Gao et~al.(2026)Gao, Zhao, An, Wu, Cao, Xiong, Huang, Wang, Yang, Su,
  Wang, Qu, Zheng, and Wang]{RollPacker}
Gao, W., Zhao, Y., An, D., Wu, T., Cao, L., Xiong, S., Huang, J., Wang, W.,
  Yang, S., Su, W., Wang, J., Qu, L., Zheng, B., and Wang, W.
\newblock {RollPacker: Taming Long-Tail Rollouts for RL Post-Training with Tail
  Batching}.
\newblock In \emph{23rd USENIX Symposium on Networked Systems Design and
  Implementation (NSDI 26)}, pp.\  849--866, Renton, WA, May 2026. USENIX
  Association.
\newblock ISBN 978-1-939133-54-0.
\newblock URL
  \url{https://www.usenix.org/conference/nsdi26/presentation/gao-wei}.

\bibitem[Guo et~al.(2025)Guo, Yang, Zhang, Song, Wang, Zhu, Xu, Zhang, Ma, Bi,
  Zhang, Yu, Wu, Wu, Gou, Shao, Li, Gao, Liu, Xue, Wang, Wu, Feng, Lu, Zhao,
  Deng, Ruan, Dai, Chen, Ji, Li, Lin, Dai, Luo, Hao, Chen, Li, Zhang, Xu, Ding,
  Gao, Qu, Li, Guo, Li, Chen, Yuan, Tu, Qiu, Li, Cai, Ni, Liang, Chen, Dong,
  Hu, You, Gao, Guan, Huang, Yu, Wang, Zhang, Zhao, Wang, Zhang, Xu, Xia,
  Zhang, Zhang, Tang, Zhou, Li, Wang, Li, Tian, Huang, Zhang, Wang, Chen, Du,
  Ge, Zhang, Pan, Wang, Chen, Jin, Chen, Lu, Zhou, Chen, Ye, Wang, Yu, Zhou,
  Pan, Li, Zhou, Wu, Yun, Pei, Sun, Wang, Zeng, Liu, Liang, Gao, Yu, Zhang,
  Xiao, An, Liu, Wang, Chen, Nie, Cheng, Liu, Xie, Liu, Yang, Li, Su, Lin, Li,
  Jin, Shen, Chen, Sun, Wang, Song, Zhou, Wang, Shan, Li, Wang, Wei, Zhang, Xu,
  Li, Zhao, Sun, Wang, Yu, Zhang, Shi, Xiong, He, Piao, Wang, Tan, Ma, Liu,
  Guo, Ou, Wang, Gong, Zou, He, Xiong, Luo, You, Liu, Zhou, Zhu, Huang, Li,
  Zheng, Zhu, Ma, Tang, Zha, Yan, Ren, Ren, Sha, Fu, Xu, Xie, Zhang, Hao, Ma,
  Yan, Wu, Gu, Zhu, Liu, Li, Xie, Song, Pan, Huang, Xu, Zhang, and
  Zhang]{DeepseekR1}
Guo, D., Yang, D., Zhang, H., Song, J., Wang, P., Zhu, Q., Xu, R., Zhang, R.,
  Ma, S., Bi, X., Zhang, X., Yu, X., Wu, Y., Wu, Z.~F., Gou, Z., Shao, Z., Li,
  Z., Gao, Z., Liu, A., Xue, B., Wang, B., Wu, B., Feng, B., Lu, C., Zhao, C.,
  Deng, C., Ruan, C., Dai, D., Chen, D., Ji, D., Li, E., Lin, F., Dai, F., Luo,
  F., Hao, G., Chen, G., Li, G., Zhang, H., Xu, H., Ding, H., Gao, H., Qu, H.,
  Li, H., Guo, J., Li, J., Chen, J., Yuan, J., Tu, J., Qiu, J., Li, J., Cai,
  J.~L., Ni, J., Liang, J., Chen, J., Dong, K., Hu, K., You, K., Gao, K., Guan,
  K., Huang, K., Yu, K., Wang, L., Zhang, L., Zhao, L., Wang, L., Zhang, L.,
  Xu, L., Xia, L., Zhang, M., Zhang, M., Tang, M., Zhou, M., Li, M., Wang, M.,
  Li, M., Tian, N., Huang, P., Zhang, P., Wang, Q., Chen, Q., Du, Q., Ge, R.,
  Zhang, R., Pan, R., Wang, R., Chen, R.~J., Jin, R.~L., Chen, R., Lu, S.,
  Zhou, S., Chen, S., Ye, S., Wang, S., Yu, S., Zhou, S., Pan, S., Li, S.~S.,
  Zhou, S., Wu, S., Yun, T., Pei, T., Sun, T., Wang, T., Zeng, W., Liu, W.,
  Liang, W., Gao, W., Yu, W., Zhang, W., Xiao, W.~L., An, W., Liu, X., Wang,
  X., Chen, X., Nie, X., Cheng, X., Liu, X., Xie, X., Liu, X., Yang, X., Li,
  X., Su, X., Lin, X., Li, X.~Q., Jin, X., Shen, X., Chen, X., Sun, X., Wang,
  X., Song, X., Zhou, X., Wang, X., Shan, X., Li, Y.~K., Wang, Y.~Q., Wei,
  Y.~X., Zhang, Y., Xu, Y., Li, Y., Zhao, Y., Sun, Y., Wang, Y., Yu, Y., Zhang,
  Y., Shi, Y., Xiong, Y., He, Y., Piao, Y., Wang, Y., Tan, Y., Ma, Y., Liu, Y.,
  Guo, Y., Ou, Y., Wang, Y., Gong, Y., Zou, Y., He, Y., Xiong, Y., Luo, Y.,
  You, Y., Liu, Y., Zhou, Y., Zhu, Y.~X., Huang, Y., Li, Y., Zheng, Y., Zhu,
  Y., Ma, Y., Tang, Y., Zha, Y., Yan, Y., Ren, Z.~Z., Ren, Z., Sha, Z., Fu, Z.,
  Xu, Z., Xie, Z., Zhang, Z., Hao, Z., Ma, Z., Yan, Z., Wu, Z., Gu, Z., Zhu,
  Z., Liu, Z., Li, Z., Xie, Z., Song, Z., Pan, Z., Huang, Z., Xu, Z., Zhang,
  Z., and Zhang, Z.
\newblock {DeepSeek-R1 Incentivizes Reasoning in LLMs through Reinforcement
  Learning}.
\newblock \emph{Nature}, 645\penalty0 (8081):\penalty0 633–638, Sept 2025.
\newblock ISSN 1476-4687.
\newblock \doi{10.1038/s41586-025-09422-z}.
\newblock URL \url{http://dx.doi.org/10.1038/s41586-025-09422-z}.

\bibitem[He et~al.(2026)He, Li, Feng, Du, Liu, Liu, Xia, and Chen]{RhymeRL}
He, J., Li, T., Feng, E., Du, D., Liu, Q., Liu, T., Xia, Y., and Chen, H.
\newblock History doesn't repeat itself but rollouts rhyme: Accelerating
  reinforcement learning with rhymerl.
\newblock In \emph{Proceedings of the 31st ACM International Conference on
  Architectural Support for Programming Languages and Operating Systems, Volume
  2}, ASPLOS '26, pp.\  929–945, New York, NY, USA, 2026. Association for
  Computing Machinery.
\newblock ISBN 9798400723599.
\newblock \doi{10.1145/3779212.3790172}.
\newblock URL \url{https://doi.org/10.1145/3779212.3790172}.

\bibitem[Hu et~al.(2026)Hu, Yang, Guo, Yao, Lin, Gu, Cai, Gan, Klimovic, and
  Han]{hu2026tlt}
Hu, Q., Yang, S., Guo, J., Yao, X., Lin, Y., Gu, Y., Cai, H., Gan, C.,
  Klimovic, A., and Han, S.
\newblock {Taming the Long-Tail: Efficient Reasoning RL Training with Adaptive
  Drafter}.
\newblock In \emph{Proceedings of the 31st ACM International Conference on
  Architectural Support for Programming Languages and Operating Systems, Volume
  2}, ASPLOS '26. Association for Computing Machinery, 2026.
\newblock \doi{10.1145/3779212.3790231}.

\bibitem[Kang et~al.(2026)Kang, Kang, Suresh, Kim, Pimpalgaonkar, Hwang, and
  Mirhoseini]{ThunderSyncRL}
Kang, S., Kang, H., Suresh, T., Kim, Y., Pimpalgaonkar, S., Hwang, S.~J., and
  Mirhoseini, A.
\newblock {ThunderSyncRL: Lossless Acceleration of Agentic Reinforcement
  Learning}, 2026.
\newblock URL \url{https://arxiv.org/abs/2610.05935}.

\bibitem[Kingma \& Ba(2017)Kingma and Ba]{AdaM}
Kingma, D.~P. and Ba, J.
\newblock {Adam: A Method for Stochastic Optimization}, 2017.
\newblock URL \url{https://arxiv.org/abs/1412.6980}.

\bibitem[Krell et~al.(2021)Krell, Kosec, Perez, and Fitzgibbon]{packing}
Krell, M.~M., Kosec, M., Perez, S.~P., and Fitzgibbon, A.
\newblock {Efficient Sequence Packing without Cross-Contamination: Accelerating
  Large Language Models without Impacting Performance}, 2021.

\bibitem[Kwon et~al.(2023)Kwon, Li, Zhuang, Sheng, Zheng, Yu, Gonzalez, Zhang,
  and Stoica]{vllm}
Kwon, W., Li, Z., Zhuang, S., Sheng, Y., Zheng, L., Yu, C.~H., Gonzalez, J.,
  Zhang, H., and Stoica, I.
\newblock {Efficient Memory Management for Large Language Model Serving with
  PagedAttention}.
\newblock In \emph{Proceedings of the 29th Symposium on Operating Systems
  Principles}, SOSP '23, pp.\  611–626, New York, NY, USA, 2023. Association
  for Computing Machinery.
\newblock ISBN 9798400702297.
\newblock \doi{10.1145/3600006.3613165}.
\newblock URL \url{https://doi.org/10.1145/3600006.3613165}.

\bibitem[Liu et~al.(2025)Liu, Zhu, Hegde, Cao, Yuan, Suwito, Griggs, Zaharia,
  Gonzalez, and Stoica]{skyrlsql}
Liu, S., Zhu, A., Hegde, S., Cao, S., Yuan, S., Suwito, S., Griggs, T.,
  Zaharia, M., Gonzalez, J.~E., and Stoica, I.
\newblock {SkyRL-SQL: Multi-Turn SQL Data Agents via RL}.
\newblock In \emph{First Workshop on Multi-Turn Interactions in Large Language
  Models}, 2025.
\newblock URL \url{https://openreview.net/forum?id=VNPGUGbC1p}.

\bibitem[Moritz et~al.(2018)Moritz, Nishihara, Wang, Tumanov, Liaw, Liang,
  Elibol, Yang, Paul, Jordan, and Stoica]{Ray}
Moritz, P., Nishihara, R., Wang, S., Tumanov, A., Liaw, R., Liang, E., Elibol,
  M., Yang, Z., Paul, W., Jordan, M.~I., and Stoica, I.
\newblock {Ray: A Distributed Framework for Emerging AI Applications}.
\newblock In \emph{13th USENIX Symposium on Operating Systems Design and
  Implementation (OSDI 18)}, pp.\  561--577, Carlsbad, CA, October 2018. USENIX
  Association.
\newblock ISBN 978-1-939133-08-3.
\newblock URL
  \url{https://www.usenix.org/conference/osdi18/presentation/moritz}.

\bibitem[{NVIDIA}(2026{\natexlab{a}})]{NVL72_GB200}
{NVIDIA}.
\newblock {NVIDIA GB200 NVL72}.
\newblock \url{https://www.nvidia.com/en-us/data-center/gb200-nvl72/}, August
  2026{\natexlab{a}}.

\bibitem[{NVIDIA}(2026{\natexlab{b}})]{NVL72_GB300}
{NVIDIA}.
\newblock {NVIDIA GB300 NVL72}.
\newblock \url{https://www.nvidia.com/en-us/data-center/gb300-nvl72/}, July
  2026{\natexlab{b}}.

\bibitem[Olmo et~al.(2026)Olmo, :, Ettinger, Bertsch, Kuehl, Graham, Heineman,
  Groeneveld, Brahman, Timbers, Ivison, Morrison, Poznanski, Lo, Soldaini,
  Jordan, Chen, Noukhovitch, Lambert, Walsh, Dasigi, Berry, Malik, Shah, Geng,
  Arora, Gupta, Anderson, Xiao, Murray, Romero, Graf, Asai, Bhagia, Wettig,
  Liu, Rangapur, Anastasiades, Huang, Schwenk, Trivedi, Magnusson, Lochner,
  Liu, Miranda, Sap, Morgan, Schmitz, Guerquin, Wilson, Huff, Bras, Xin, Shao,
  Skjonsberg, Shen, Li, Wilde, Pyatkin, Merrill, Chang, Gu, Zeng, Sabharwal,
  Zettlemoyer, Koh, Farhadi, Smith, and Hajishirzi]{olmo3}
Olmo, T., :, Ettinger, A., Bertsch, A., Kuehl, B., Graham, D., Heineman, D.,
  Groeneveld, D., Brahman, F., Timbers, F., Ivison, H., Morrison, J.,
  Poznanski, J., Lo, K., Soldaini, L., Jordan, M., Chen, M., Noukhovitch, M.,
  Lambert, N., Walsh, P., Dasigi, P., Berry, R., Malik, S., Shah, S., Geng, S.,
  Arora, S., Gupta, S., Anderson, T., Xiao, T., Murray, T., Romero, T., Graf,
  V., Asai, A., Bhagia, A., Wettig, A., Liu, A., Rangapur, A., Anastasiades,
  C., Huang, C., Schwenk, D., Trivedi, H., Magnusson, I., Lochner, J., Liu, J.,
  Miranda, L. J.~V., Sap, M., Morgan, M., Schmitz, M., Guerquin, M., Wilson,
  M., Huff, R., Bras, R.~L., Xin, R., Shao, R., Skjonsberg, S., Shen, S.~Z.,
  Li, S.~S., Wilde, T., Pyatkin, V., Merrill, W., Chang, Y., Gu, Y., Zeng, Z.,
  Sabharwal, A., Zettlemoyer, L., Koh, P.~W., Farhadi, A., Smith, N.~A., and
  Hajishirzi, H.
\newblock {Olmo 3}, 2026.
\newblock URL \url{https://arxiv.org/abs/2512.13961}.

\bibitem[OpenAI et~al.(2026)OpenAI, :, Jaech, Kalai, Lerer, Richardson,
  El-Kishky, Low, Helyar, Madry, Beutel, Carney, Iftimie, Karpenko, Passos,
  Neitz, Prokofiev, Wei, Tam, Bennett, Kumar, Saraiva, Vallone, Duberstein,
  Kondrich, Mishchenko, Applebaum, Jiang, Nair, Zoph, Ghorbani, Zhang, Rossen,
  Sokolowsky, Barak, McGrew, Minaiev, Hao, Baker, Houghton, McKinzie, Eastman,
  Lugaresi, Bassin, Hudson, Li, de~Bourcy, Voss, Shen, Zhang, Koch, Orsinger,
  Hesse, Fischer, Chan, Roberts, Kappler, Levy, Selsam, Dohan, Farhi, Mely,
  Robinson, Tsipras, Li, Oprica, Freeman, Zhang, Wong, Proehl, Cheung,
  Mitchell, Wallace, Ritter, Mays, Wang, Such, Raso, Leoni, Tsimpourlas, Song,
  von Lohmann, Sulit, Salmon, Parascandolo, Chabot, Zhao, Brockman, Leclerc,
  Salman, Bao, Sheng, Andrin, Bagherinezhad, Ren, Lightman, Chung, Kivlichan,
  O'Connell, Osband, Gilaberte, Akkaya, Kostrikov, Sutskever, Kofman, Pachocki,
  Lennon, Wei, Harb, Twore, Feng, Yu, Weng, Tang, Yu, Candela, Palermo, Parish,
  Heidecke, Hallman, Rizzo, Gordon, Uesato, Ward, Huizinga, Wang, Chen, Xiao,
  Singhal, Nguyen, Cobbe, Shi, Wood, Rimbach, Gu-Lemberg, Liu, Lu, Stone, Yu,
  Ahmad, Yang, Liu, Maksin, Ho, Fedus, Weng, Li, McCallum, Held, Kuhn,
  Kondraciuk, Kaiser, Metz, Boyd, Trebacz, Joglekar, Chen, Tintor, Meyer,
  Jones, Kaufer, Schwarzer, Shah, Yatbaz, Guan, Xu, Yan, Glaese, Chen, Lampe,
  Malek, Wang, Fradin, McClay, Pavlov, Wang, Wang, Murati, Bavarian,
  Rohaninejad, McAleese, Chowdhury, Chowdhury, Ryder, Tezak, Brown, Nachum,
  Boiko, Murk, Watkins, Chao, Ashbourne, Izmailov, Zhokhov, Dias, Arora, Lin,
  Lopes, Gaon, Miyara, Leike, Hwang, Garg, Brown, James, Shu, Cheu, Greene,
  Jain, Altman, Toizer, Toyer, Miserendino, Agarwal, Hernandez, Baker,
  McKinney, Yan, Zhao, Hu, Santurkar, Chaudhuri, Zhang, Fu, Papay, Lin, Balaji,
  Sanjeev, Sidor, Broda, Clark, Wang, Gordon, Sanders, Patwardhan, Sottiaux,
  Degry, Dimson, Zheng, Garipov, Stasi, Bansal, Creech, Peterson, Eloundou, Qi,
  Kosaraju, Monaco, Pong, Fomenko, Zheng, Zhou, Zhan, McCabe, Zaremba, Dubois,
  Lu, Chen, Cha, Bai, He, Zhang, Wang, Shao, and Li]{openaio1}
OpenAI, :, Jaech, A., Kalai, A., Lerer, A., Richardson, A., El-Kishky, A., Low,
  A., Helyar, A., Madry, A., Beutel, A., Carney, A., Iftimie, A., Karpenko, A.,
  Passos, A.~T., Neitz, A., Prokofiev, A., Wei, A., Tam, A., Bennett, A.,
  Kumar, A., Saraiva, A., Vallone, A., Duberstein, A., Kondrich, A.,
  Mishchenko, A., Applebaum, A., Jiang, A., Nair, A., Zoph, B., Ghorbani, B.,
  Zhang, B., Rossen, B., Sokolowsky, B., Barak, B., McGrew, B., Minaiev, B.,
  Hao, B., Baker, B., Houghton, B., McKinzie, B., Eastman, B., Lugaresi, C.,
  Bassin, C., Hudson, C., Li, C.~M., de~Bourcy, C., Voss, C., Shen, C., Zhang,
  C., Koch, C., Orsinger, C., Hesse, C., Fischer, C., Chan, C., Roberts, D.,
  Kappler, D., Levy, D., Selsam, D., Dohan, D., Farhi, D., Mely, D., Robinson,
  D., Tsipras, D., Li, D., Oprica, D., Freeman, E., Zhang, E., Wong, E.,
  Proehl, E., Cheung, E., Mitchell, E., Wallace, E., Ritter, E., Mays, E.,
  Wang, F., Such, F.~P., Raso, F., Leoni, F., Tsimpourlas, F., Song, F., von
  Lohmann, F., Sulit, F., Salmon, G., Parascandolo, G., Chabot, G., Zhao, G.,
  Brockman, G., Leclerc, G., Salman, H., Bao, H., Sheng, H., Andrin, H.,
  Bagherinezhad, H., Ren, H., Lightman, H., Chung, H.~W., Kivlichan, I.,
  O'Connell, I., Osband, I., Gilaberte, I.~C., Akkaya, I., Kostrikov, I.,
  Sutskever, I., Kofman, I., Pachocki, J., Lennon, J., Wei, J., Harb, J.,
  Twore, J., Feng, J., Yu, J., Weng, J., Tang, J., Yu, J., Candela, J.~Q.,
  Palermo, J., Parish, J., Heidecke, J., Hallman, J., Rizzo, J., Gordon, J.,
  Uesato, J., Ward, J., Huizinga, J., Wang, J., Chen, K., Xiao, K., Singhal,
  K., Nguyen, K., Cobbe, K., Shi, K., Wood, K., Rimbach, K., Gu-Lemberg, K.,
  Liu, K., Lu, K., Stone, K., Yu, K., Ahmad, L., Yang, L., Liu, L., Maksin, L.,
  Ho, L., Fedus, L., Weng, L., Li, L., McCallum, L., Held, L., Kuhn, L.,
  Kondraciuk, L., Kaiser, L., Metz, L., Boyd, M., Trebacz, M., Joglekar, M.,
  Chen, M., Tintor, M., Meyer, M., Jones, M., Kaufer, M., Schwarzer, M., Shah,
  M., Yatbaz, M., Guan, M.~Y., Xu, M., Yan, M., Glaese, M., Chen, M., Lampe,
  M., Malek, M., Wang, M., Fradin, M., McClay, M., Pavlov, M., Wang, M., Wang,
  M., Murati, M., Bavarian, M., Rohaninejad, M., McAleese, N., Chowdhury, N.,
  Chowdhury, N., Ryder, N., Tezak, N., Brown, N., Nachum, O., Boiko, O., Murk,
  O., Watkins, O., Chao, P., Ashbourne, P., Izmailov, P., Zhokhov, P., Dias,
  R., Arora, R., Lin, R., Lopes, R.~G., Gaon, R., Miyara, R., Leike, R., Hwang,
  R., Garg, R., Brown, R., James, R., Shu, R., Cheu, R., Greene, R., Jain, S.,
  Altman, S., Toizer, S., Toyer, S., Miserendino, S., Agarwal, S., Hernandez,
  S., Baker, S., McKinney, S., Yan, S., Zhao, S., Hu, S., Santurkar, S.,
  Chaudhuri, S.~R., Zhang, S., Fu, S., Papay, S., Lin, S., Balaji, S., Sanjeev,
  S., Sidor, S., Broda, T., Clark, A., Wang, T., Gordon, T., Sanders, T.,
  Patwardhan, T., Sottiaux, T., Degry, T., Dimson, T., Zheng, T., Garipov, T.,
  Stasi, T., Bansal, T., Creech, T., Peterson, T., Eloundou, T., Qi, V.,
  Kosaraju, V., Monaco, V., Pong, V., Fomenko, V., Zheng, W., Zhou, W., Zhan,
  W., McCabe, W., Zaremba, W., Dubois, Y., Lu, Y., Chen, Y., Cha, Y., Bai, Y.,
  He, Y., Zhang, Y., Wang, Y., Shao, Z., and Li, Z.
\newblock {OpenAI o1 System Card}, 2026.
\newblock URL \url{https://arxiv.org/abs/2412.16720}.

\bibitem[Ren et~al.(2021)Ren, Rajbhandari, Aminabadi, Ruwase, Yang, Zhang, Li,
  and He]{zerooffload}
Ren, J., Rajbhandari, S., Aminabadi, R.~Y., Ruwase, O., Yang, S., Zhang, M.,
  Li, D., and He, Y.
\newblock {ZeRO-Offload: Democratizing Billion-Scale Model Training}, 2021.
\newblock URL \url{https://arxiv.org/abs/2101.06840}.

\bibitem[Schulman et~al.(2017)Schulman, Wolski, Dhariwal, Radford, and
  Klimov]{PPO}
Schulman, J., Wolski, F., Dhariwal, P., Radford, A., and Klimov, O.
\newblock {Proximal Policy Optimization Algorithms}, 2017.
\newblock URL \url{https://arxiv.org/abs/1707.06347}.

\bibitem[Shao et~al.(2024)Shao, Wang, Zhu, Xu, Song, Bi, Zhang, Zhang, Li, Wu,
  and Guo]{GRPO}
Shao, Z., Wang, P., Zhu, Q., Xu, R., Song, J., Bi, X., Zhang, H., Zhang, M.,
  Li, Y.~K., Wu, Y., and Guo, D.
\newblock {DeepSeekMath: Pushing the Limits of Mathematical Reasoning in Open
  Language Models}.
\newblock \emph{arXiv preprint arXiv:2402.03300}, 2024.
\newblock \doi{10.48550/arXiv.2402.03300}.

\bibitem[Shao et~al.(2026)Shao, Srivatsa, Srivastava, Wu, Ariyak, Wu, Patel,
  Wang, Liang, Dao, Zhang, Zhang, Athiwaratkun, Xu, and
  Wang]{shao2026beatlongtail}
Shao, Z., Srivatsa, V., Srivastava, S., Wu, Q., Ariyak, A., Wu, X., Patel, A.,
  Wang, J., Liang, P., Dao, T., Zhang, C., Zhang, Y., Athiwaratkun, B., Xu, C.,
  and Wang, J.
\newblock {Beat the Long Tail: Distribution-Aware Speculative Decoding for RL
  Training}.
\newblock In \emph{Proceedings of the 9th Annual Conference on Machine Learning
  and Systems}, MLSys '26, 2026.
\newblock \doi{10.48550/arXiv.2511.13841}.
\newblock arXiv:2511.13841.

\bibitem[Sheng et~al.(2025)Sheng, Zhang, Ye, Wu, Zhang, Zhang, Peng, Lin, and
  Wu]{verl}
Sheng, G., Zhang, C., Ye, Z., Wu, X., Zhang, W., Zhang, R., Peng, Y., Lin, H.,
  and Wu, C.
\newblock {HybridFlow: A Flexible and Efficient RLHF Framework}.
\newblock In \emph{Proceedings of the Twentieth European Conference on Computer
  Systems}, pp.\  1279--1297, 2025.

\bibitem[Shoeybi et~al.(2020)Shoeybi, Patwary, Puri, LeGresley, Casper, and
  Catanzaro]{Megatron}
Shoeybi, M., Patwary, M., Puri, R., LeGresley, P., Casper, J., and Catanzaro,
  B.
\newblock {Megatron-LM: Training Multi-Billion Parameter Language Models Using
  Model Parallelism}, 2020.
\newblock URL \url{https://arxiv.org/abs/1909.08053}.

\bibitem[Wang et~al.(2026)Wang, Zhou, Chen, Zhang, Zhang, Luo, Yin, Wu, Jiang,
  Yao, Zheng, and Zhang]{QwenGyre}
Wang, W., Zhou, Y., Chen, M., Zhang, S., Zhang, Y., Luo, Y., Yin, Z., Wu, C.,
  Jiang, J., Yao, W., Zheng, C., and Zhang, J.
\newblock {QwenGyre: An Elastic Reinforcement Learning Framework for Training
  xLong-Horizon Agents}, 2026.
\newblock URL \url{https://arxiv.org/abs/2609.33848}.

\bibitem[Yao et~al.(2023)Yao, Zhao, Yu, Du, Shafran, Narasimhan, and
  Cao]{react}
Yao, S., Zhao, J., Yu, D., Du, N., Shafran, I., Narasimhan, K., and Cao, Y.
\newblock {ReAct: Synergizing Reasoning and Acting in Language Models}, 2023.
\newblock URL \url{https://arxiv.org/abs/2210.03629}.

\bibitem[Yeung \& Gardner(2026)Yeung and Gardner]{Cerebras}
Yeung, A. and Gardner, E.
\newblock {Introducing Cerebras CS-4: The Fastest AI Just Got Faster}.
\newblock Cerebras Systems Blog,
  \url{https://www.cerebras.ai/blog/introducing-cerebras-cs-4}, August 2026.

\bibitem[Yu et~al.(2025)Yu, Zhang, Zhu, Yuan, Zuo, Yue, Dai, Fan, Liu, Liu,
  Liu, Lin, Lin, Ma, Sheng, Tong, Zhang, Zhang, Zhang, Zhu, Zhu, Chen, Chen,
  Wang, Yu, Song, Wei, Zhou, Liu, Ma, Zhang, Yan, Qiao, Wu, and Wang]{DAPO}
Yu, Q., Zhang, Z., Zhu, R., Yuan, Y., Zuo, X., Yue, Y., Dai, W., Fan, T., Liu,
  G., Liu, L., Liu, X., Lin, H., Lin, Z., Ma, B., Sheng, G., Tong, Y., Zhang,
  C., Zhang, M., Zhang, W., Zhu, H., Zhu, J., Chen, J., Chen, J., Wang, C., Yu,
  H., Song, Y., Wei, X., Zhou, H., Liu, J., Ma, W.-Y., Zhang, Y.-Q., Yan, L.,
  Qiao, M., Wu, Y., and Wang, M.
\newblock {DAPO: An Open-Source LLM Reinforcement Learning System at Scale},
  2025.
\newblock URL \url{https://arxiv.org/abs/2503.14476}.

\bibitem[Zhao et~al.(2023)Zhao, Gu, Varma, Luo, Huang, Xu, Wright, Shojanazeri,
  Ott, Shleifer, Desmaison, Balioglu, Damania, Nguyen, Chauhan, Hao, Mathews,
  and Li]{fsdp}
Zhao, Y., Gu, A., Varma, R., Luo, L., Huang, C.-C., Xu, M., Wright, L.,
  Shojanazeri, H., Ott, M., Shleifer, S., Desmaison, A., Balioglu, C., Damania,
  P., Nguyen, B., Chauhan, G., Hao, Y., Mathews, A., and Li, S.
\newblock {PyTorch FSDP: Experiences on Scaling Fully Sharded Data Parallel},
  2023.
\newblock URL \url{https://arxiv.org/abs/2304.11277}.

\bibitem[Zheng et~al.(2024)Zheng, Yin, Xie, Sun, Huang, Yu, Cao, Kozyrakis,
  Stoica, Gonzalez, Barrett, and Sheng]{sglang}
Zheng, L., Yin, L., Xie, Z., Sun, C., Huang, J., Yu, C.~H., Cao, S., Kozyrakis,
  C., Stoica, I., Gonzalez, J.~E., Barrett, C., and Sheng, Y.
\newblock {SGLang: Efficient Execution of Structured Language Model Programs}.
\newblock In \emph{Proceedings of the 38th International Conference on Neural
  Information Processing Systems}, NeurIPS '24, Red Hook, NY, USA, 2024. Curran
  Associates Inc.
\newblock ISBN 9798331314385.

\bibitem[Zhong et~al.(2025{\natexlab{a}})Zhong, Zhang, Song, Hu, Jin, Wu, Chen,
  Chen, Zhou, Wan, Zhou, Jiang, Zhu, and Jiang]{StreamRL}
Zhong, Y., Zhang, Z., Song, X., Hu, H., Jin, C., Wu, B., Chen, N., Chen, Y.,
  Zhou, Y., Wan, C., Zhou, H., Jiang, Y., Zhu, Y., and Jiang, D.
\newblock {StreamRL: Scalable, Heterogeneous, and Elastic RL for LLMs with
  Disaggregated Stream Generation}, 2025{\natexlab{a}}.
\newblock URL \url{https://arxiv.org/abs/2504.15930}.

\bibitem[Zhong et~al.(2025{\natexlab{b}})Zhong, Zhang, Wu, Liu, Chen, Wan, Hu,
  Xia, Ming, Zhu, and Jin]{RLHFuse}
Zhong, Y., Zhang, Z., Wu, B., Liu, S., Chen, Y., Wan, C., Hu, H., Xia, L.,
  Ming, R., Zhu, Y., and Jin, X.
\newblock {Optimizing RLHF Training for Large Language Models with Stage
  Fusion}.
\newblock In \emph{Proceedings of the 22nd USENIX Symposium on Networked
  Systems Design and Implementation}, NSDI '25, USA, 2025{\natexlab{b}}. USENIX
  Association.
\newblock ISBN 978-1-939133-46-5.

\bibitem[Zhu et~al.(2026)Zhu, Gu, Sriram, Jacob, Kamahori, Li, Pan, Xie, Wang,
  Krishnamurthy, and Kasikci]{servingstudio}
Zhu, K., Gu, M., Sriram, S., Jacob, M., Kamahori, K., Li, V., Pan, Y., Xie, D.,
  Wang, S., Krishnamurthy, A., and Kasikci, B.
\newblock {Introducing ServingStudio: An Integrated Workbench for Simulating,
  Analyzing, and Optimizing LLM Serving Systems}.
\newblock
  \url{https://syfi.cs.washington.edu/blog/2026-09-24-introducing-servingstudio/},
  September 2026.
\newblock Code: \url{https://github.com/SyFI-ServingStudio/ServingStudio}.

\bibitem[Zhu et~al.(2025)Zhu, Xie, Lv, and slime Contributors]{Slime}
Zhu, Z., Xie, C., Lv, X., and slime Contributors.
\newblock {slime: An LLM Post-Training Framework for RL Scaling}.
\newblock \url{https://github.com/THUDM/slime}, 2025.
\newblock GitHub repository. Corresponding author: Xin Lv.

\end{thebibliography}
\end{document}